\documentclass[lettersize,journal]{IEEEtran}

\usepackage{epsfig}
\usepackage{graphicx}
\usepackage{amsmath}
\usepackage{amssymb}
\usepackage[citecolor=blue, colorlinks]{hyperref}
\usepackage{comment}
\usepackage{color}
\usepackage{overpic}
\usepackage{subcaption}
\usepackage{algorithm2e}
\usepackage[dvipsnames]{xcolor}
\usepackage{color}
\usepackage[switch,columnwise]{lineno}
\runningpagewiselinenumbers

\def\BibTeX{{\rm B\kern-.05em{\sc i\kern-.025em b}\kern-.08em
    T\kern-.1667em\lower.7ex\hbox{E}\kern-.125emX}}
\usepackage{balance}

\begin{document}

\title{On Optimal Sampling for Learning SDF \\ Using MLPs Equipped with Positional Encoding}

\author{IEEE Publication Technology Department
\thanks{Manuscript created October, 2020; This work was developed by the IEEE Publication Technology Department. This work is distributed under the \LaTeX \ Project Public License (LPPL) ( http://www.latex-project.org/ ) version 1.3. A copy of the LPPL, version 1.3, is included in the base \LaTeX \ documentation of all distributions of \LaTeX \ released 2003/12/01 or later. The opinions expressed here are entirely that of the author. No warranty is expressed or implied. User assumes all risk.}}

\markboth{Journal of \LaTeX\ Class Files,~Vol.~18, No.~9, September~2020}%
{How to Use the IEEEtran \LaTeX \ Templates}

\author{Guying~Lin*,
        Lei~Yang*,
        Yuan~Liu,
        Congyi~Zhang,\\
        Junhui~Hou,
        Xiaogang~Jin,
        Taku~Komura,
        John~Keyser,
        Wenping~Wang
\IEEEcompsocitemizethanks{
\IEEEcompsocthanksitem
Guying Lin, Lei Yang, Yuan Liu, and Taku Komura are with the University of Hong Kong. Email: guyinglin2000@gmail.com, lyang@cs.hku.hk, yuanly@connect.hku.hk, taku@cs.hku.hk.
\IEEEcompsocthanksitem
Congyi Zhang is with the University of British Columbia. Email: congyiz@cs.ubc.ca
\IEEEcompsocthanksitem
Junhui Hou is with the Department of Computer Science, City University of Hong Kong. Email: jh.hou@cityu.edu.hk.
\IEEEcompsocthanksitem
Xiaogang Jin is with the State Key Lab of CAD\&CG, Zhejiang University. Email: jin@cad.zju.edu.cn.
\IEEEcompsocthanksitem
John Keyser and Wenping Wang are with Texas A\&M University. Email: keyser@cse.tamu.edu, wenping@tamu.edu.
}
\thanks{*Both authors contributed equally to this research.}
}

\maketitle

\begin{abstract}
Neural implicit fields, such as the neural signed distance field (SDF) of a shape, have emerged as a powerful representation for many applications, e.g., encoding a 3D shape and performing collision detection. Typically, implicit fields are encoded by Multi-layer Perceptrons (MLP) with positional encoding (PE) to capture high-frequency geometric details. However, a notable side effect of such PE-equipped MLPs is the noisy artifacts present in the learned implicit fields. While increasing the sampling rate could in general mitigate these artifacts, in this paper we aim to explain this adverse phenomenon through the lens of Fourier analysis. We devise a tool to determine the appropriate sampling rate for learning an accurate neural implicit field without undesirable side effects.  Specifically, we propose a simple yet effective method to estimate the intrinsic frequency of a given network with randomized weights based on the Fourier analysis of the network’s responses. It is observed that a PE-equipped MLP has an intrinsic frequency much higher than the highest frequency component in the PE layer. Sampling against this intrinsic frequency following the Nyquist-Sannon sampling theorem allows us to determine an appropriate training sampling rate. We empirically show in the setting of SDF fitting that this recommended sampling rate is sufficient to secure accurate fitting results, while further increasing the sampling rate would not further noticeably reduce the fitting error. Training PE-equipped MLPs simply with our sampling strategy leads to performances superior to the existing methods.
\end{abstract}

\begin{IEEEkeywords}
SDF, neural representation, positional encoding, Fourier analysis, spectrum analysis, neural network.
\end{IEEEkeywords}

\thispagestyle{empty}
\section{Introduction}\label{sec:intro}

\IEEEPARstart{C}{oordinate}-based networks, typically MLPs (Multilayer Perceptrons) taking the coordinates of points in a low dimensional space as inputs, have emerged as a general representation of encoding implicit fields for 2D and 3D contents~\cite{mildenhall2020nerf,park2019deepsdf,gropp2020implicit,sitzmann2020implicit,atzmon2020sal,liu2020neural,martel2021acorn,wang2021neus,xie2021neural}. 
These neural implicit representations enjoy several major benefits over their traditional counterparts. They offer a compact representation since they only need to store a relatively small number of network weights, exhibit a strong capability in representing complex geometries, and ensure a smooth representation with the use of smooth nonlinear activation functions as opposed to discrete geometric representations (e.g., meshes and point clouds) for a 3D representation.

\begin{figure}[t]
    \begin{subfigure}[t]{0.3\linewidth}
      \includegraphics[width=\textwidth]{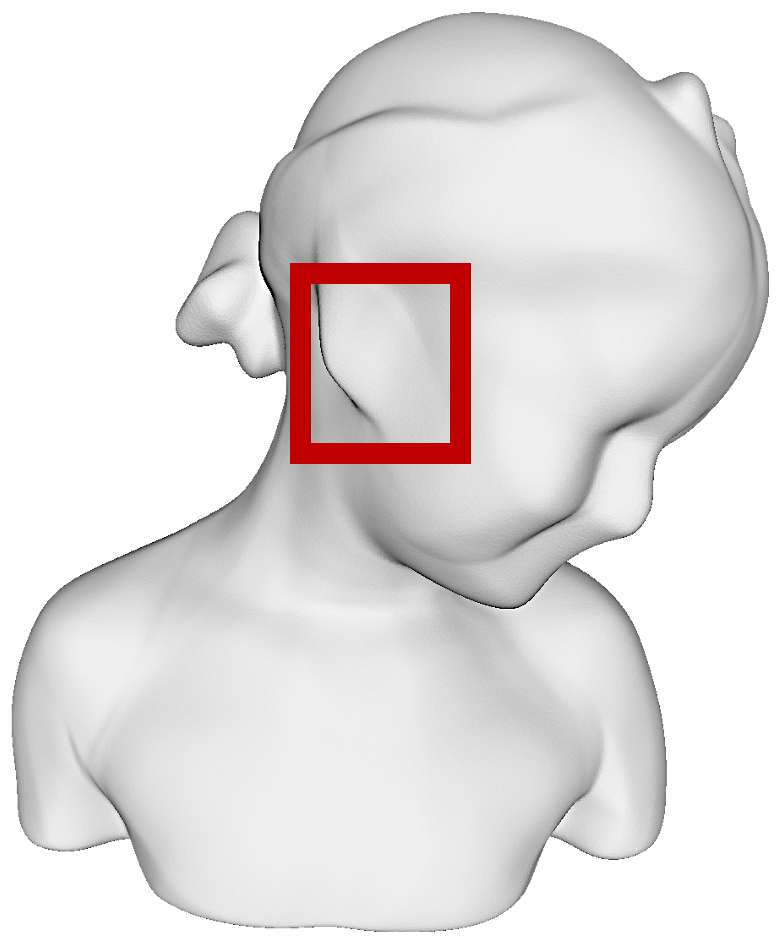}
      \caption{No PE}\label{fig:teaser_a}
    \end{subfigure}
    \hfill
    \begin{subfigure}[t]{0.3\linewidth}
    \includegraphics[width=\textwidth]{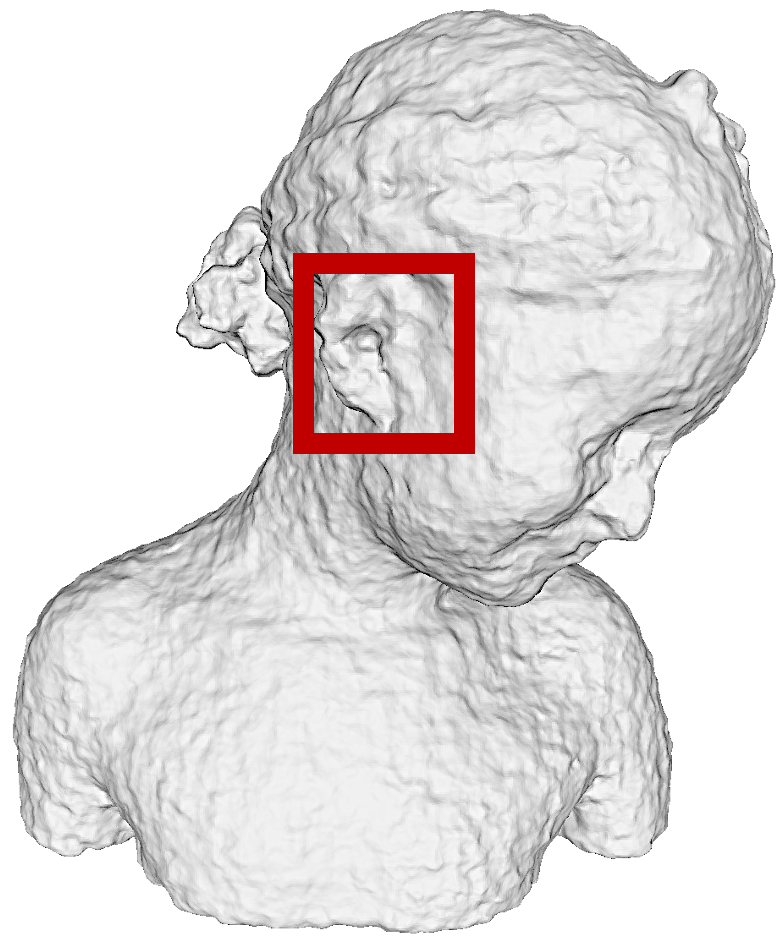}\caption{PE with insufficient sampling}\label{fig:teaser_b}
    \end{subfigure}
    \hfill
    \begin{subfigure}[t]{0.3\linewidth}
      \includegraphics[width=\textwidth]{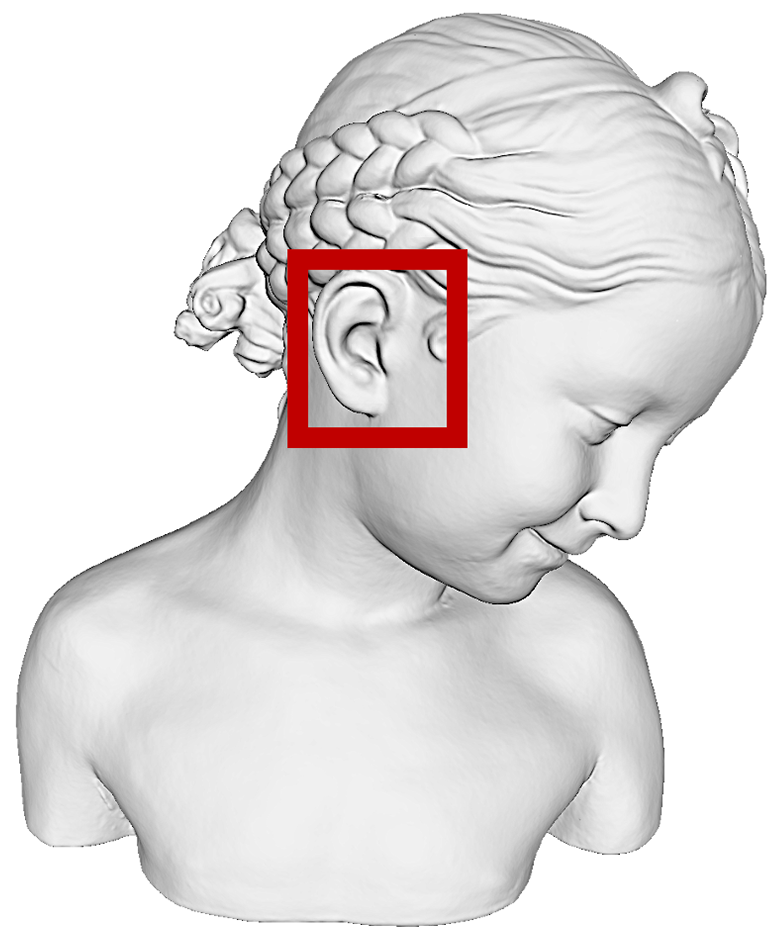}\caption{PE with our sampling}\label{fig:teaser_c}
    \end{subfigure}
    \caption{\textbf{The benefits from positional encoding (PE) and our sampling strategy.} Zero-level sets of the learned signed distance field (SDF) on the \textit{Bimba} model, using (a) vanilla MLP without PE, (b) PE-equipped MLP with insufficient training samples (33k points), and (c) PE-equipped MLP with our recommended sampling strategy, respectively. 
    }
  \end{figure}

However, an MLP alone often fails to capture high-frequency details of the target fields~\cite{tancik2020fourier,mildenhall2020nerf,sitzmann2020implicit}, which is coined as the spectral bias/frequency principle as explained by Rahaman et al.~\cite{rahaman2019spectral} or Xu et al.~\cite{xu2019frequency}. Take the task of fitting a Signed Distance Field (SDF) as an example. A single MLP network (8-layer MLP with 512 neurons per layer) usually produces over-smoothed results, as shown in Fig.~\ref{fig:teaser_a}, especially in the regions of the SDF zero-level set which contains intricate geometry details. Similar observations have also been made in tasks like learning a neural radiance field~\cite{mildenhall2020nerf}. To address this problem, the sinusoidal positional encoding, called {\em PE} for short, is introduced~\cite{mildenhall2020nerf} for enhancing the ability of the MLP to capture these geometry details. Specifically, PE adds a layer of sinusoidal functions with various frequencies to an ordinary MLP. However, na\"{i}ve application of PE in neural implicit representations often suffers from the side-effect of producing \textit{wavy artifacts} ~\cite{davies2020effectiveness,gropp2020implicit,tancik2020fourier} as shown in Fig.~\ref{fig:teaser_b}.

In this paper, we present an in-depth analysis of the cause of the wavy artifacts. 
Our analysis is grounded in the \textit{response frequency} of a neural network~\cite{xu2019frequency}, which characterizes the frequency components of the network's output. Specifically, there are very high frequencies in the response frequency spectrum that, if undersampled, can lead to aliasing effects and result in wavy artifacts.

Then, the training loss only minimizes the difference between the aliased network outputs and the target field, which leaves the underlying high-frequency components of the network outputs not suppressed,  resulting in wavy artifacts and high test error at inference time. 

\begin{figure}[t]
    \centering
    \begin{overpic}
    [width=\linewidth, clip]{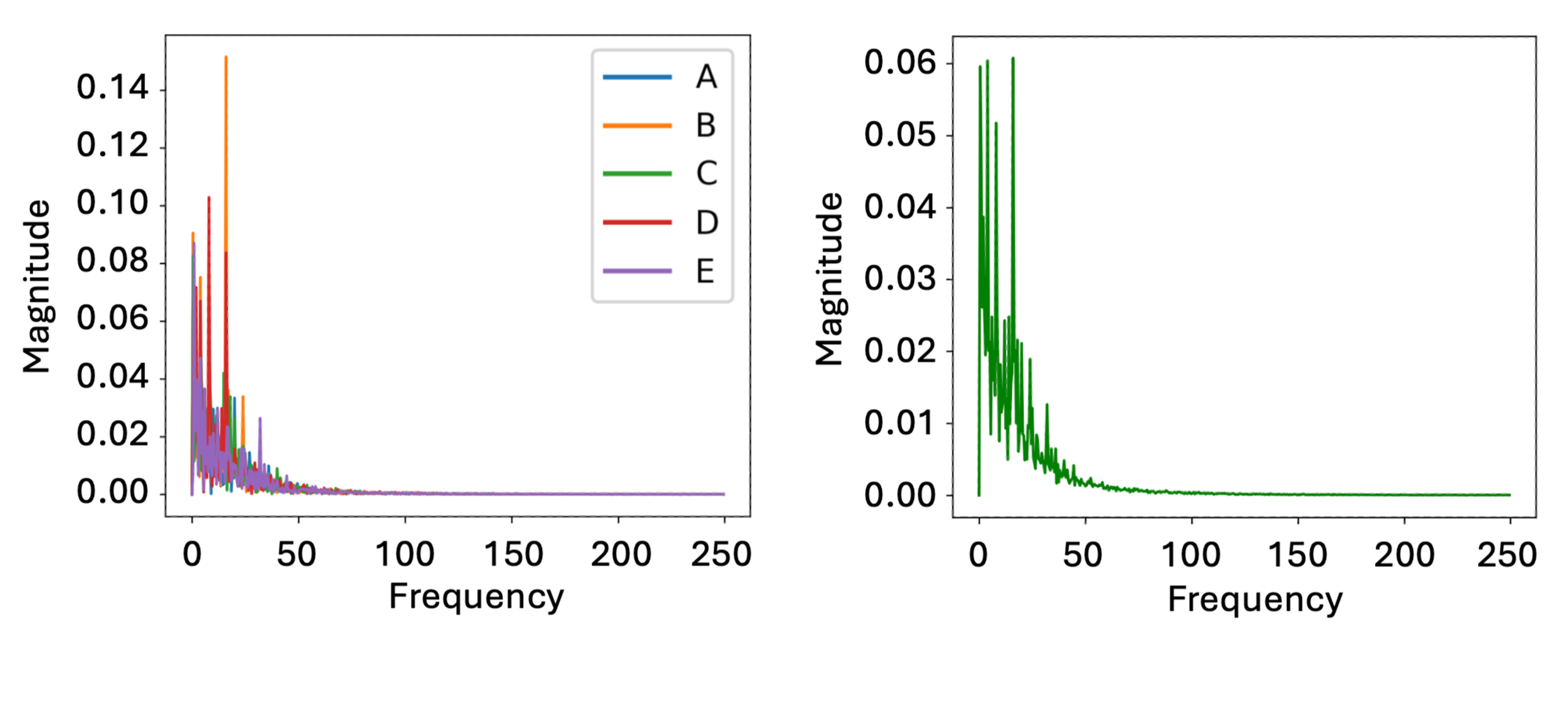}
    \put(10,0){(a) Multiple Spectra}
    \put(57,0){(b) Intrinsic Spectrum}
    \end{overpic}
    \caption{\textbf{Intrinsic frequency of a network.} (a) Consistent response frequency spectra are observed in randomized PE-equipped MLPs with the same architecture (8 layers with 512 neurons per layer); (b) The averaged spectrum of the network over multiple randomized network weights is defined as the intrinsic spectrum of the PE-equipped MLP. }
    \label{fig:intrinsic_frequency_expectation}
\end{figure}

Our first finding is that the randomly initialized PE-equipped MLP networks with the same architecture have very similar spectrum profiles to each other. 
As shown in Fig.~\ref{fig:intrinsic_frequency_expectation} (a), we randomly initialize the MLP networks 5 times with different random sets of weights and visualize their output frequency spectra. As we can see, these MLP networks all show similar spectrum profiles and bandwidths.
Based on this observation, we compute the statistical expectation of these spectra in Fig.~\ref{fig:intrinsic_frequency_expectation} (b) and term it the \textit{intrinsic spectrum} (explained in Sec.~\ref{sec:intrinsic_freq}) to these PE-equipped MLP networks with this same architecture. 

From Fig.~\ref{fig:intrinsic_frequency_expectation} (b) we can see that the frequencies of the network are predominantly distributed in the low-value region but there are always some high-frequency components (up to 60 Hz) in the tail.

We validate the observation that when the training samples are insufficient to recover these high-frequency components in the intrinsic spectrum according to the Nyquist-Shannon (NS) sampling theorem~\cite{ns_sample}, the learned SDFs often end up with wavy artifacts at test time.  We show that the wavy artifacts can be removed by using sufficient sampling on these high-frequency components. This demonstrates that the aliasing effect is the cause of the wavy artifacts at test time. 
This motivates us to identify a cut-off frequency of the intrinsic spectrum to filter out those high-frequency components of negligible energy, thus not affecting the approximation quality. 
We also show that further increasing the training sampling rate beyond this cut-off frequency brings little additional improvement in the setting of SDF fitting. This suggests an effective strategy for sampling data points for training a neural implicit representation. 

We devise an empirical method for probing the intrinsic spectrum of an MLP network and determining the cut-off frequency of this spectrum.
We show experimentally that a PE-equipped MLP trained using the appropriate sampling rate recommended by our study can produce high-quality fitting of SDF that is superior to several widely used methods like SIREN~\cite{sitzmann2020implicit}, NGLOD~\cite{takikawa2021neural}, or MLP equipped with Spline-PE~\cite{wang2021spline}. Furthermore, a comprehensive analysis is provided to validate the effectiveness of our designs in the recommendation of the highest intrinsic frequency (and thus the sampling rate) of PE-equipped MLP networks.

Our contributions are summarized as follows: 

1) We present an insightful observation that the optimal sampling rate in SDF fitting is affected by both the target SDF frequency and the response frequency of MLP networks, the latter of which is characterized by the intrinsic spectrum of the PE-equipped MLP network.  

2) We propose a practical method for estimating the intrinsic spectrum of a PE-equipped MLP and recommending the optimal sampling rate based on the cut-off frequency for the intrinsic spectrum, as suggested by the Nyquist-Shannon sampling theorem. This optimal sampling rate helps remove wavy artifacts to produce high-quality learned SDF fitting.

3) We provide extensive quantitative comparisons with state-of-the-art methods for SDF fitting and show the superiority of a PE-equipped MLP as an accurate 3D representation if trained properly with a sufficient data sampling rate.

\section{Related works}\label{sec:related_work}

Recent studies~\cite{mescheder2019occupancy,chen2019learning,park2019deepsdf,michalkiewicz2019implicit,sitzmann2020metasdf,tancik2021learned,atzmon2020sal,atzmon2020sald,gropp2020implicit,davies2020effectiveness,chibane2020neural,lipman2021phase,wang2021spline,sitzmann2020implicit,Williams2021CVPR} have demonstrated that it is promising to represent 3D shapes or scenes as implicit functions (e.g., signed distance functions or occupancy fields) parameterized by a deep neural network. 
These recent efforts show that a well-trained network can drastically compress the memory footprint to several megabytes for representing 3D shapes. 
To attain a high-fidelity neural representation, especially well reconstructing the geometric details in the given shape, many techniques have been developed, such as positional encoding (PE) techniques that map the input to a higher dimensional vector~\cite{mildenhall2020nerf,tancik2020fourier,wang2021spline} or specially designed activation functions~\cite{sitzmann2020implicit,ramasinghe2021beyond}.

In this work, we consider fitting an MLP network with sinusoidal PE~\cite{mildenhall2020nerf} to represent a 3D shape. In particular, we explore how spatial samples and sinusoidal PE (or PE for short) can work together to enable MLP networks to capture fine-scale geometric features. We compare our method with several representative methods, e.g.,~\cite{sitzmann2020implicit, wang2021spline}.

Another line of work~\cite{peng2020convolutional,chibane2020implicit,jiang2020local,martel2021acorn,takikawa2021neural,tang2021octfield,muller2022instant} leverages the learned latent features for enhancing the expressiveness on fine-scale details. 
In particular, these studies usually learn a shallow MLP network and a set of latent feature vectors, each associated with a spatial coordinate, forming an \textit{explicit} grid map. Given a spatial query, the corresponding latent feature vector is obtained, such as by interpolating nearby latent feature vectors stored in the grid map. Then, this latent feature vector is mapped by the MLP head to the value of the implicit function at this spatial query. While the results are compelling, this is achieved at the expense of explicitly storing a grid feature map. Artifacts at the interface of two adjacent grids may be observed as well. 

Among them, NGLOD~\cite{takikawa2021neural} is proposed to represent a single complex 3D shape as the signed distance function induced by the shape using a hierarchically organized feature map. It achieves the SOTA performance in both surface reconstruction and SDF approximation. We will validate our design choices by comparing our method with NGLOD.

\textbf{Spectral bias of the MLP networks.} \cite{rahaman2019spectral} and~\cite{xu2019frequency} unveiled the characteristics of an MLP that learns low-frequency signals of the dataset first and gradually fits the high-frequency components. This characteristic is coined as Spectral bias~\cite{rahaman2019spectral} or Frequency principle~\cite{xu2019frequency}. \cite{jacot2018neural} approach this problem from the Neural Tangent Kernel viewpoint. \cite{tancik2020fourier} then applied this explanation to the training dynamics of neural implicit representation, answering why low-frequency contents are learned in the first place. However, this line of work emphasizes the learning dynamics of the MLPs or neural implicit representations, while offering limited insight into why positional encoding schemes, in general, will lead to noisy outputs.

A recent work~\cite{ramasinghe2022frequency} demonstrated, via a Fourier lens, that the noisy outputs are due to the spectral energy shift from the low-frequency end to a high-frequency range when positional encoding layers are added. This observation, along with its analysis based on a two-layer network, conforms to ours in the frequency probing of various 1D examples and 3D SDF learning tasks. In contrast to~\cite{ramasinghe2022frequency} where a regularization term is proposed to incorporate a smooth geometric prior for training, we first discuss in depth that the \textit{cause} of the noisy outputs is due to the aliasing effect during training. Then, we propose a simple yet effective sampling strategy to overcome the aliasing effect that ensures satisfactory results at inference time.

We also observed works on leveraging non-Fourier positional encoding schemes~\cite{ramasinghe2023learnable} or incorporating bandlimited filters~\cite{lindell2022bacon} for image fitting tasks. Our work is complementary to these recent studies, revisiting the sinusoidal positional encoding scheme and unveiling the reason for its widely observed artifacts. Our outcome provides a practical choice for practitioners when using sinusoidal positional encoding to train their neural networks~\cite{koptev2022neural,wong2022learning}. 

\section{Methodology}\label{sec:methods}

\subsection{Learning SDF with Sinusoidal PE}

\begin{figure}[t]
    \centering
    \begin{overpic}
    [width=\linewidth]{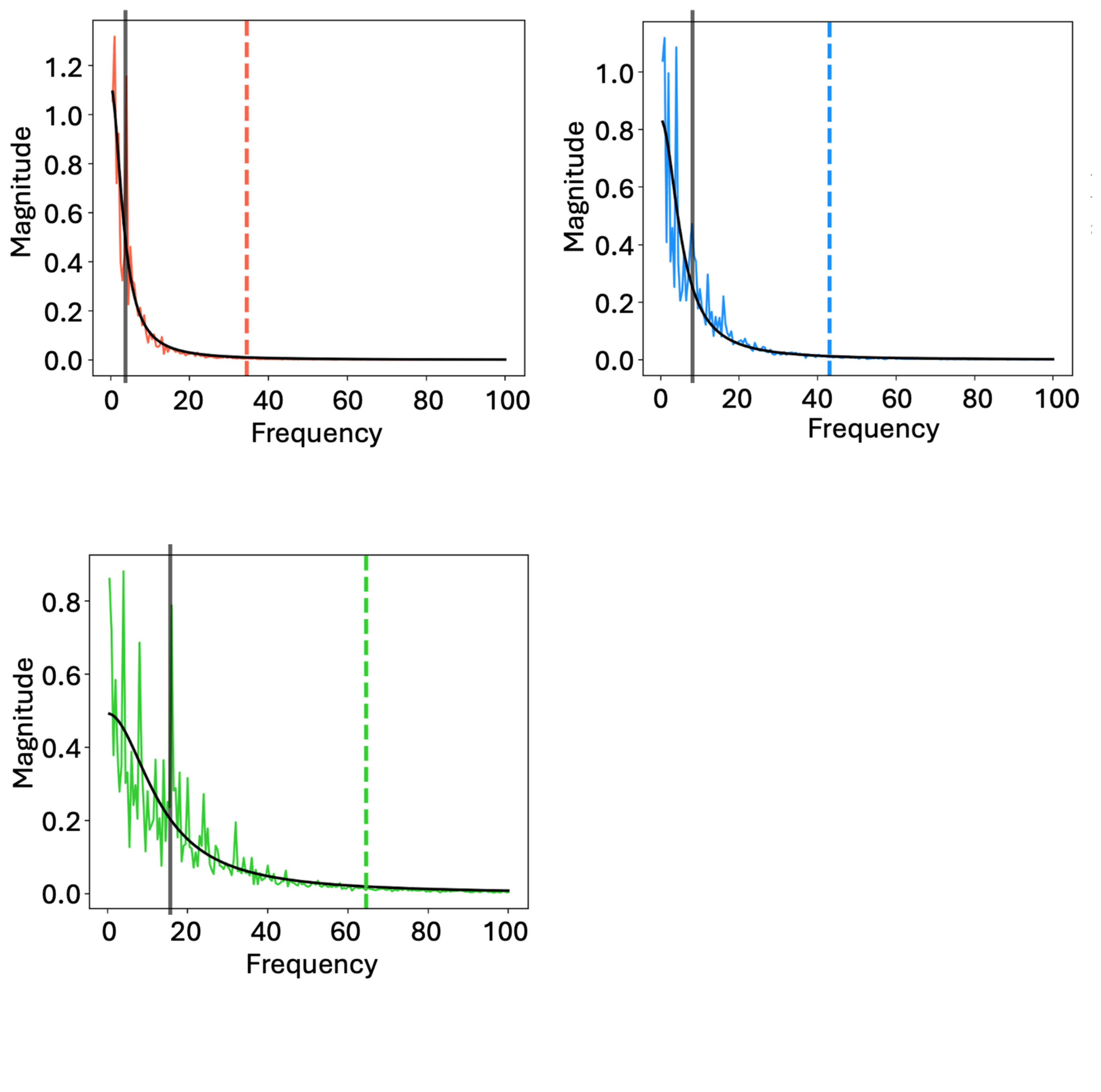}
    \put(20,54){(a) D = 3}
    \put(70,54){(b) D = 4}
    \put(20,6){(c) D = 5}

    \end{overpic}
    \caption{\textbf{Intrinsic spectra of PE-equipped MLPs with different $D$}. The degree of sinusoidal positional encoding (PE), $D$, increases from 3 to 5. The highest frequency in the sinusoidal PE is marked with the vertical black line. The networks carry high-frequency components with large magnitudes beyond the PE's highest frequency. The cut-off frequency of the networks (vertical dashed lines) is defined to determine the sampling rate and avoid aliasing.
    The magnitude beyond the cut-off frequency exhibits negligible values.}
    \label{fig:d1_dft}
\end{figure}

A {\em signed distance field} (SDF) returns the signed distance of a query point $\mathbf{x} \in \mathbb{R}^3$ to a given surface in 3D space. We follow the convention of defining the sign to be positive if the query point is outside the shape and negative otherwise; therefore the surface is the zero-level set of the SDF. One can encode a signed distance function as an MLP network denoted $f_{\theta}(\cdot)$, parameterized by the network weights.
It has been shown that equipping an MLP network with sinusoidal positional encoding~\cite{mildenhall2020nerf,tancik2020fourier} can improve the network's representation ability for representing complex shapes. Such a PE-equipped MLP can be written as a composite function
\begin{equation}\label{eq:pe}
    y = f_{\theta} ( \gamma_D(\mathbf{x}) ),
\end{equation}
Here, the sinusoidal positional encoding is represented as 
$$\gamma_D(\mathbf{x}) = [\mathbf{x}, PE_0(\mathbf{x}), PE_1(\mathbf{x}), \cdot\cdot\cdot, PE_D(\mathbf{x})],$$
where $PE_p(x) = [\sin(2^p \pi x), \cos(2^p \pi x)]$ and $x$ is one of the components of $\mathbf{x}$. 
The highest frequency of the combined signal of the PE encoding is $2^{(D-1)}$. $D$ is termed as the degree of the sinusoidal PE.

The following SDF loss is adopted in fitting an SDF field with the neural network,
\begin{align}\label{eq:objective_sdf}
    E = \int_{\Omega } |SDF(\mathbf{x}) - f_{\theta}(\gamma_D(\mathbf{x}))| d\mathbf{x},
\end{align}
where $\Omega$ is the spatial domain and $SDF(\mathbf{x})$ is the ground-truth (GT) SDF at $\mathbf{x}$. While incorporating a PE layer with high-frequency components benefits the fitting of high-frequency details, it also brings a side effect of noises~\cite{tancik2020fourier} or wavy artifacts~\cite{davies2020effectiveness} as shown in Fig.~\ref{fig:teaser_b}.

\textbf{Overview. }
It has been empirically observed that increasing the sampling rate can mitigate the noisy artifacts of the SDF encoded by a PE-equipped MLP. In this work, we take a step further by first investigating the relevant mechanism of the network to understand the cause of this side effect of adding PE, and then estimating quantitatively the sufficient sampling rate for suppressing these artifacts.
We show that this side effect of adding PE can be viewed as an aliasing problem. In particular, we observe that this aliasing problem is caused during the minimization of Eq.~\ref{eq:objective_sdf} where the loss is optimized concerning a set of aliased frequency components in $f_{\theta}(\gamma_D(\mathbf{x}))$, the output of the network, due to an insufficient sampling rate.

In the following, we will provide an intuitive analysis of $f_{\theta}(\gamma_D(\mathbf{x}))$ and the potential aliasing effect at training that explains the wavy artifacts. 
Then, we propose a practical solution to efficiently assess the intrinsic spectrum of a PE-equipped MLP and an effective method for determining the cut-off frequency of the intrinsic spectrum of the MLP to obtain the optimal sampling rate for training the MLP. Finally, we devise a simple sampling and training strategy to achieve high-quality SDF fitting using a PE-equipped MLP.

\subsection{Training Aliasing Effect}

An aliasing problem is caused by the failure of sampling signal components with frequencies higher than the sampling rate. According to the Nyquist-Shannon (NS) sampling theorem~\cite{ns_sample}, at least a uniform sampling rate of $2F$ is required to reconstruct the given signal with frequency $F$. 

Let $f_{\theta}(\cdot)$ denote an MLP without the PE layer, parameterized by the weights and biases $\theta=\{\mathbf{W}_i, \mathbf{b}_i\}_{i=1}^L$.  Let $\gamma(\mathbf{x})$ denote the PE layer, which has no free parameters to optimize. 
Then we denote the PE-equipped MLP by
\[
h_{PE}(x) \equiv f_{\theta}(\gamma(\mathbf{x})).
\]
The optimization problem of fitting an implicit field $\bar{y}(\mathbf{x})$ is achieved by minimizing the following loss function
\begin{equation} 
    E(\mathbf{x}) = \arg\min_{\theta} |h_{PE}(\mathbf{x}) - \bar{y}(\mathbf{x})|. 
\end{equation}

Optimizing the network parameters $\theta$ requires a sufficiently dense sampling of the function $h_{PE}(\mathbf{x})$, which is based on the highest frequency of $h_{PE}(\mathbf{x})$, as suggested by the Nyquist-Shannon sampling theorem. Note that $h_{PE}(\mathbf{x}) \equiv f_{\theta}(\gamma(\mathbf{x}))$ usually carries components of much higher frequency than those of the PE signal $\gamma(\mathbf{x})$ due to the action of multiple layers of the MLP $f_{\theta}(\cdot)$ applied to $\gamma(\mathbf{x})$. 

When an insufficient sampling rate with respect to $h_{PE}(\mathbf{x})$ is applied for training $h_{PE}(\mathbf{x})$, it will result in an aliased function of lower frequency, denoted $h_{PE}^{lf}(\mathbf{x})$. Conceptually, we have the decomposition
\[
h_{PE} (\mathbf{x}) = h_{PE}^{lf}(\mathbf{x}) + h_{PE}^{hf}(\mathbf{x})
\]
where $h_{PE}^{hf}(\mathbf{x})$ contains the components of $h_{PE} (\mathbf{x})$ that are of higher frequency than those contained in the aliased function $h_{PE}^{lf}(\mathbf{x})$.

Hence, the optimization problem under this under-sampling condition turns out to be minimizing an aliased target function $h_{PE}^{lf}(\mathbf{x})$ as follows
\begin{equation}\label{eq:fit_aliased_function}
    \tilde{E}(x)=\arg\min_{\theta} | h_{PE}^{lf}(\mathbf{x}) - \bar{y}(\mathbf{x}) |.
\end{equation}
This leaves the higher-frequency components of $h_{PE}^{hf}(\mathbf{x})$ untrained. 
Consequently, given a query $\mathbf{x}$ unseen at the training stage, the fitting error $E(\mathbf{x})$ generated by the high-frequency components $h_{PE}^{hf}(\mathbf{x})$ will be present in the final output of the PE-equipped network $h_{PE} (\mathbf{x})$ at inference time, leading to wavy artifacts of the SDF approximation.

From this analysis, it is clear that an insufficient number of training samples will lead to wavy artifacts at inference time due to the aliasing effect during training. Hence, to determine the sufficient sampling rate, we need a method for examining the frequency characteristics of $f_{\theta}(\gamma_D(\mathbf{x}))$ and determining the appropriate sampling frequency.

\subsection{Intrinsic Spectrum and Cut-off Frequency}
\label{sec:intrinsic_freq}

\textbf{Intrinsic spectrum. }
Due to the complicated interactions between the layer inputs and the interleaved linear/non-linear operations, it is difficult, if not intractable, to theoretically analyze the frequency characteristics intrinsic to a network architecture $f_{\theta}(\gamma(x))$. Therefore, we devise an empirical approach to probing the frequency characteristics by studying specific randomized networks. It is observed that the frequency spectra of these networks with randomized weights have similar spectra.

Given a randomized network $f_{\theta}$, we first use densely distributed regular points $\mathbf{x}_j$ to produce network output $y_j$, from which we obtain the frequency spectrum $Y(F)$ by applying fast Fourier transform (FFT) to $y_j$. 
Then, we define the intrinsic spectrum as follows:
\begin{equation}\label{eq:intrinsic_spectrum}
    \mathbb{E}(F) = \sum_{i=1}^M Y_i(F) / M
\end{equation}
In our implementation, each output signal of such a network is produced with a set of densely equidistant sampled points sampled on a line through the bounding domain of the network input.
Dense points ensure that FFT captures any high-frequency components.

As we can see from Fig.~\ref{fig:intrinsic_frequency_expectation} (a), when a network is assigned different sets of random weights, which are drawn from the same distribution, their frequency spectra are similar to each other, showing a limited bandwidth within a relatively tight range towards the low-frequency end. Furthermore, as shown in Fig.~\ref{fig:intrinsic_frequency_expectation} (b), the fact that the expectation of the frequency spectra possesses this bandlimited property is intrinsic to the PE-equipped networks initialized from the same distribution. Therefore, this intrinsic property observed from the frequency spectrum allows us to estimate a cut-off frequency for determining the appropriate training sampling rate to mitigate the aliasing effect at inference time. In Fig.~\ref{fig:d1_dft}, we show the intrinsic spectra of PE-equipped MLP with different frequency levels $D$ of PE.

\textbf{Cut-off frequency and sampling rate.}
Next, we define a \textit{cut-off frequency} for this frequency spectrum to characterize the undersampling and oversampling conditions. The sampling rate corresponding to the cut-off frequency ensures the suppression of the wavy artifacts in the learned SDF by avoiding the undersampling or training aliasing issue. Meanwhile, we show that using a large sampling rate beyond this cut-off frequency will not noticeably further minimize the SDF fitting error for the given MLP network. Thus, knowing the cut-off frequency helps prevent oversampling to avoid wasting computing resources. 

We derive the cut-off frequency as follows. We fit a smooth curve $C(F)$ to the intrinsic spectrum $\mathbb{E}(F)$ as follows:

\begin{equation}\label{eq:fit_spectrum}
    C(F) = \frac{a}{{\mathbb{E}(F)^2 + b}},
\end{equation}
where $a$ and $b$ are the coefficients for fitting.

\begin{figure}
    \centering
    \begin{overpic}
    [width=\linewidth, clip]{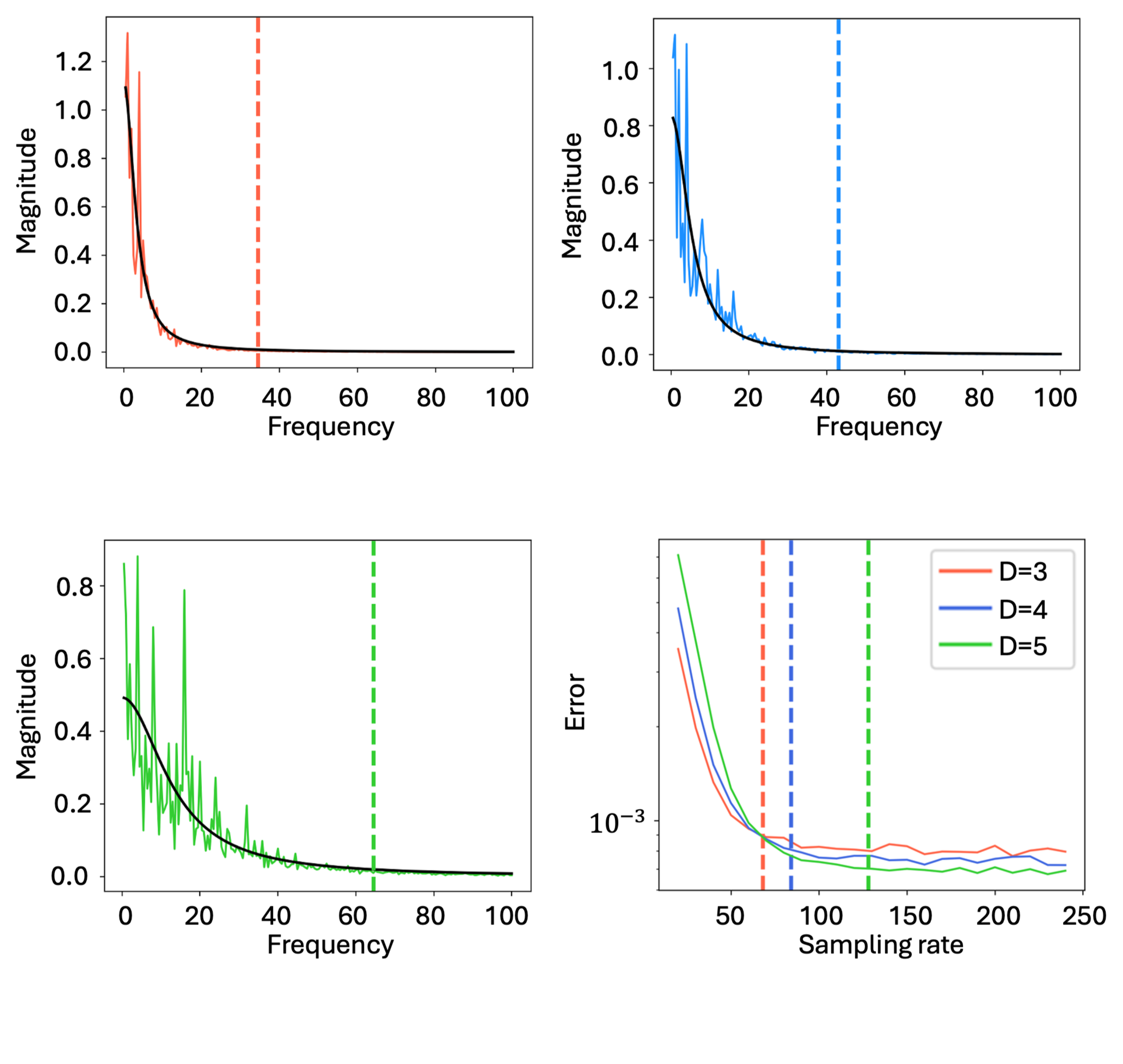}
    \put(10,50){(a) Spectrum (D=3)}
    \put(61,50){(c) Spectrum (D=4)}
    \put(10,4){(d) Spectrum (D=5)}
    \put(63,4){(d) SDF error}

    \end{overpic}
    \caption{\textbf{Validation of the recommended sampling rate.} (a) to (c) show the spectra of PE-equipped MLPs (8 layers with 512 neurons per layer) using different $D$ (the degree of PE). Black curves are fit to the spectra for determining the respective cut-off frequencies (vertical dashed lines).
    SDF convergence errors against the sampling rates are shown in (d).
    }
    \label{fig:sample_error}
    \vspace{-2mm}
\end{figure}

The fit curves are shown as the black curves in Fig.~\ref{fig:sample_error} (a-c). 
To find the cut-off frequency at which the spectrum is almost constant, we compute the derivative of $C(F)$, that is $C(F)'$. We empirically set the cut-off frequency $F_c$ as the lowest frequency whose $C(F_c)' = 6 \times 10^{-4}$. This hyperparameter has been found to work well in all of our experiments.

Our recommended sampling rate corresponding to the cut-off frequency $F_c$ is twice $F_c$ according to the NS sampling theorem. We adopt the multi-dimensional generalization of the Nyquist-Shannon sampling theorem~\cite{ns} as follows:
\begin{equation}
    \mu = (2 F_{c})^\ell,
    \label{eq:nq}
\end{equation}
where $\ell$ indicates the input dimension of the task.

The process of finding the cut-off frequency of a randomized MLP network is summarized in Alg.~\ref{alg:alg_density}.
\RestyleAlgo{ruled}
\SetKwComment{Comment}{/* }{ */}
\begin{algorithm}
\caption{Determining the sampling density.}\label{alg:alg_density}
\KwData{Neural Network $f(\cdot | \theta_i)$; Task dimension $\ell$.}
\KwResult{Suggested sample density $\mu$.}
1) Given several neural network $f(\cdot | \theta_i)$ (randomly initialized)\;
2) Densely sample a set of equidistant points $\mathbf{x}_j$ along a specified axis
\Comment*[r]{The number of samples should be sufficiently large }
3) Obtain network output/response $y^i_j = f( \gamma_D(\mathbf{x}_j) | \theta_i)$\;
4) Obtain response frequency spectrum $Y_i(F) = \mathrm{FFT}(y^i_j)$ 
\Comment*[r]{FFT stands for Fast Fourier Transform}
5) Compute the intrinsic spectrum as Eq.~\ref{eq:intrinsic_spectrum}\;
6) Fit $C(F)$ to intrinsic spectrum $\mathbb{E}(F)$ using Eq.~\ref{eq:fit_spectrum}\\
7) Compute derivative $C(F)'$ with respect to $F$\;
8) Cut-off frequency $F_c$ computed as 
$$C(F_c)' = 6 \times 10^{-4}$$\\
9) Suggested sampling density
\[\mu = (2 F_{c})^\ell\]
\end{algorithm}

\section{Experiments and results}\label{sec:results}

In this section, we describe the evaluation datasets and metrics in Sec.~\ref{sec:protocol}. 
Experiments are conducted in the following aspects to demonstrate the effectiveness of our method. We begin with the validation of the proposed sampling rate in Sec.~\ref{sec:exp_sample_rate}. Then, we compare our results with those produced by existing methods on the SDF fitting task in Sec.~\ref{sec:results}. Finally, comprehensive analyses of our method, including surface fitting quality and its applicability to various different network initializations and architectures, are provided in Sec.~\ref{sec:analysis}.

\begin{figure}
    \centering
    \begin{overpic}
    [width=\linewidth, clip]{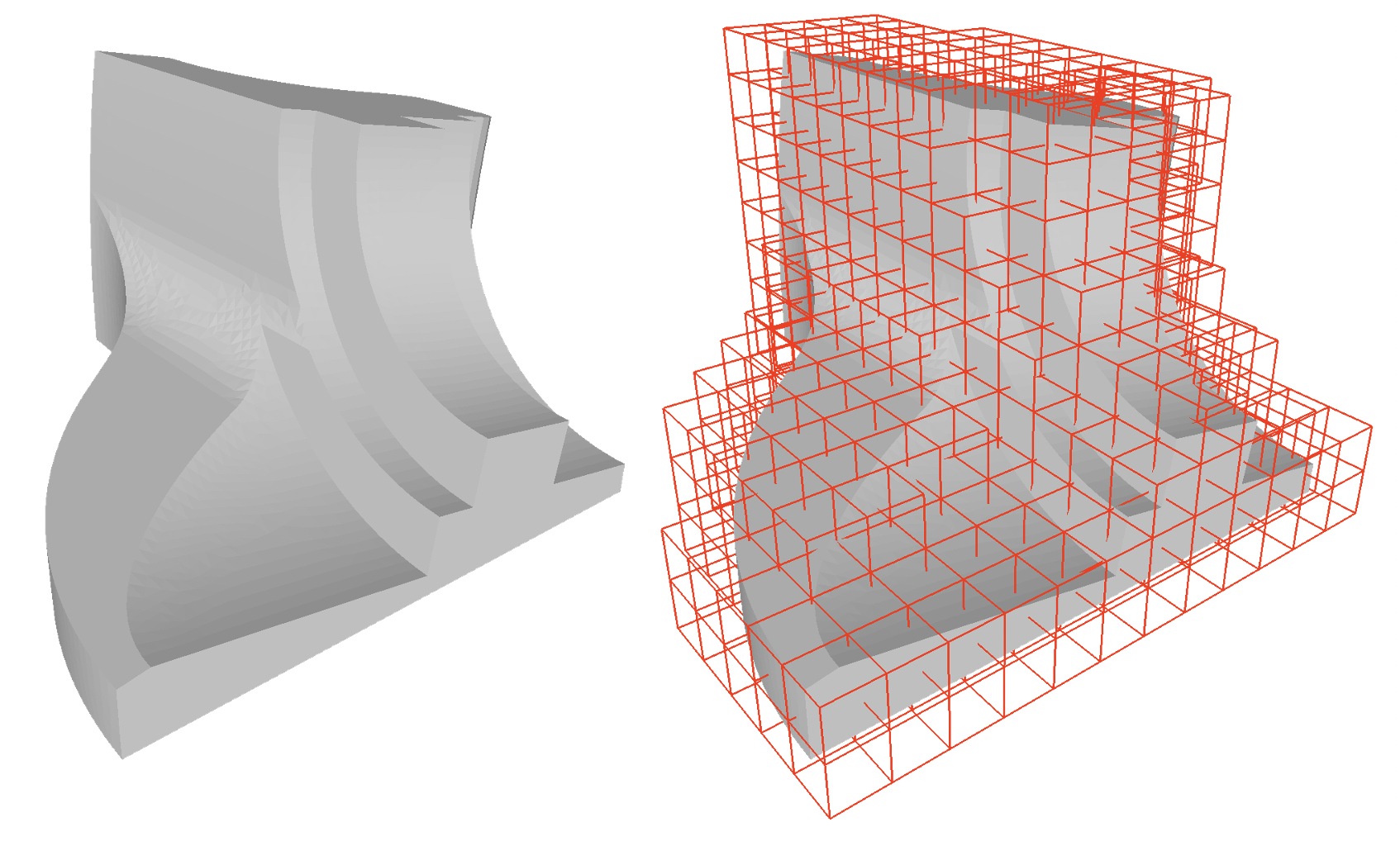}
    \put(15,0){(a) GT shape}
    \put(65,0){(b) Selected grid cells}
    \end{overpic}
    \caption{Training sample points are from the grid cells intersecting with the surface shape. 
    }
    \label{fig:active_grids}
    \vspace{-2mm}
\end{figure}

\begin{figure*}
    \centering
    \begin{overpic}
    [width=\linewidth, clip]{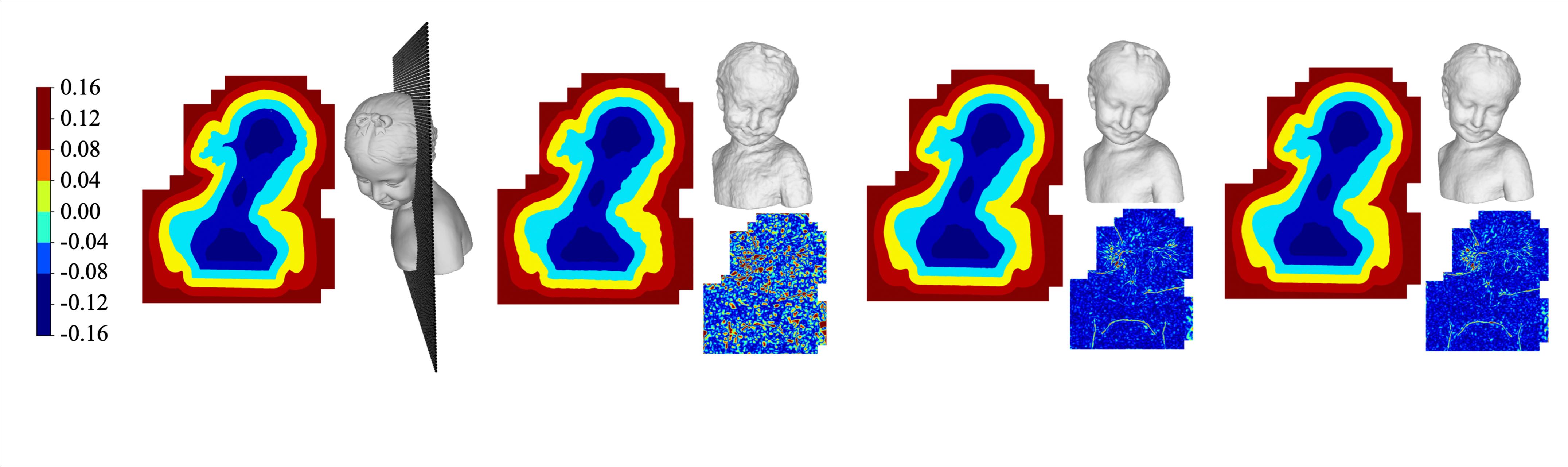}
    \put(10,1){(a) GT}
    \put(31,1){(b) PE sampling}
    \put(55,1){(c) Our sampling}
    \put(78,1){(d) Denser sampling}
    \put(30,6.5){Avg. SDF error: }
    \put(33,4){29.3E-4}
    \put(54,6.5){Avg. SDF error: }
    \put(58,4){5.6E-4}
    \put(77,6.5){Avg. SDF error:}
    \put(80,4){5.5E-4}
    \end{overpic}
    \caption{\textbf{SDF fitting results of different sampling rates.} 
    (a) Ground-truth SDF (left) on the cutting plane and the surface shape (right). 
    (b-d) show the learned SDF on the same cutting plane (left), the reconstruction, and the error map (bottom right). PE frequency with $D$=5 is employed which corresponds to a cut-off frequency of 63$Hz$.
    (b) uses $\sim$20k samples based on the highest PE frequency $D$; (c) uses $\sim$900k samples based on the cut-off frequency; (d) uses $\sim$1800k samples.
    ``Our sampling'' (c) can effectively mitigate noisy artifacts in the reconstructed surface and the learned SDF observed in (b). Comparison between (c) and (d) shows that an increase beyond the recommended sample rate brings little improvement to the fitting quality.
    }
    \label{fig:sdf_fitting_progress}
    \vspace{0mm}
\end{figure*}
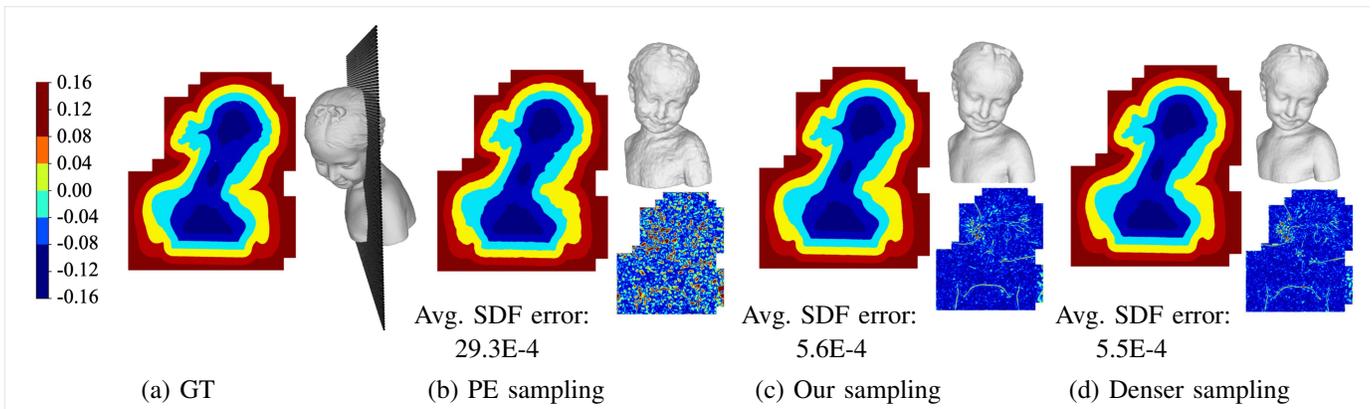

\subsection{Implementation details}\label{sec:protocol}

\textbf{Normalizing network outputs}. Before computing the frequency spectrum of the signals $\{\mathbf{x}_j, y_j\}$, we first whiten the output $y_j$ by subtracting the mean and scaling them so that their standard deviation equals 1. Note that these operations are linear and used to normalize the total energy the sequential signals $\{\mathbf{x}_j, y_j\}$ carry; they will not modify the frequency characteristic of the signal concerned. This way, the sequential data obtained from different randomized networks are now normalized and comparable to each other.

\textbf{Training details.}
Given the surface of a 3D shape, we first normalize it to a cubical domain with its maximal extent being $[-1, 1]$. 

Given a PE-MLP network, we calculate the recommended sampling rate $\mu$ as described and generate a set of grid samples covering the cubical domain based on $\mu$.
In our experiments, the spatial region around the surface shape is concerned with ensuring fair comparison with the baseline methods. Therefore, we keep only the sample points that are close to the surface shape. To this end, the entire cubical domain is partitioned into a coarse, regular grid of $20\times20\times20$ cells. The sample points that are located in the cells intersecting with the surface shape are kept, denoted $X$. We visualize these cells in Fig.~\ref{fig:active_grids}. For these sample points in $X$, we compute their ground truth SDF values to train the neural implicit shapes.

By default, we use an 8-layer MLP network equipped with $\gamma_{D=5}$. Each layer of the MLP has 512 neurons and is followed by the \textit{softplus} activation ($\beta=100$ as suggested in~\cite{atzmon2020sald}) with an exception at the last layer where a \textit{tanh} activation is applied. 
During training, we employ a mini-batch of $100,000$ sample points. 
Each network is trained for $30,000$ iterations, which takes approximately 30 mins in total time. The ADAM optimizer~\cite{kingma2015adam} is used for training. The training starts with a learning rate of $0.0001$ and then decays by a factor of 0.1 after $27,000$ iterations. 
We observed that all networks converged under this setting.
All networks are implemented with PyTorch~\cite{NEURIPS2019_9015} and trained on an NVIDIA RTX3090 graphic card (24 GB memory).

\textbf{Evaluation protocols.}
To evaluate the quality of the SDF fitting results, we adopt the mean absolute SDF error, $\epsilon_{SDF}$, between the predicted signed distance values and the corresponding ground-truth (GT) values at validation points. 
The validation set consists of $100,000$ points which are sampled in the active grid cells without overlaps with the training samples.

For evaluating the surface reconstruction quality, we use the Marching Cubes~\cite{lorensen1987marching} to extract the reconstructed surface at the zero-level set from the learned signed distance field.
Our dataset comprises eight commonly employed shapes in computer graphics. 

\subsection{Validation of our sampling rate}
\label{sec:exp_sample_rate}

To validate the effectiveness of the proposed sampling rate as described in Section~\ref{sec:intrinsic_freq}, 
we present the results on the \textit{Lucy} shape for illustrative purposes in Fig.~\ref{fig:sample_error}, and results on another two shapes are presented in the Appendix, where we observe the same trend between the sampling rate and the SDF error on these shapes. The setting of PE-MLP networks follows the default setting except that different highest frequencies $D$ are used for the PE layer.
 
Fig.~\ref{fig:sample_error} (a) to (c) 
illustrate the frequency spectrum of the output signal (or the response frequency) of the same PE-equipped MLP with a different set of randomized weights and biases. 
Each dashed vertical line represents the sampling rate ($2F$) corresponding to cut-off frequency $F$ for a PE-equipped MLP with a different $D$ value in its PE layer. 

Remarkably, each dashed vertical line in Fig.~\ref{fig:sample_error} (d) corresponding to the cut-off frequency precisely captures the point of test error convergence for each PE-equipped MLP, indicating that further increasing the sampling rate will not significantly further enhance SDF accuracy, thus validating our design choice.

We further provide qualitative results of SDF fitting in Fig.~\ref{fig:sdf_fitting_progress} 
Here, the same network architecture is used which has a PE layer of $D=5$ and eight hidden layers, each with 512 neurons.
The three learned SDFs shown in this figure are trained with different amounts of training samples: one using the sampling rate corresponding to the highest PE frequency $D$ (b), one with the sampling rate dictated by the network cut-off frequency (c), and another with a higher sampling rate (d). 
Notably, the SDF from the sampling rate based on the highest PE frequency exhibits undulating artifacts with a mean absolute error of 29.3E-4, whereas the errors of the suggested or even denser sampling rates have an error of 5.6E-4 and of 5.5E-4, respectively, which are similar to each other but significantly lower than the results with PE sampling rate. 

The results demonstrate that our recommended sampling rate serves as a sufficient and cost-effective lower bound of sample point numbers for high-fidelity SDF reconstruction. 

\subsection{Comparison with existing methods}

We compare our methods with three baseline approaches, i.e., SPE~\cite{wang2021spline}, SIREN~\cite{sitzmann2020implicit}, and NGLOD~\cite{takikawa2021neural}. 
SPE~\cite{wang2021spline} proposes an alternative positional encoding scheme based on the B-spline basis function. SIREN~\cite{sitzmann2020implicit} proposes to use a \textit{sine} activation function to replace the softplus activation function used in ours and SPE's networks and avoid employing any positional encoding schemes. 
NGLOD~\cite{takikawa2021neural} is a representative method of another line of work that makes use of explicit, learnable feature grids organized in an octree, along with a compact shared MLP network to fit the SDF. 

\textbf{Implementation of baseline methods}. 
We use the proposed approach to probe the intrinsic frequency spectrum of SPE and SIREN networks. We prepare training samples for SPE based on its recommended sampling rate, which is denser than that for our PE-MLP under the same setting. Since the cut-off frequency and thus the sampling rate of SIREN is lower than ours, we trained SIREN with the same sample points as ours to ensure fair comparison.
We adopt their default learning rate, $1\times 10^{-4}$ for SPE and $0.5\times 10^{-4}$ for SIREN, for training. Our method, SPE, and SIREN are trained for $30,000$ iterations and observed to have fully converged. 
Since NGLOD learns a hierarchical feature grid different from ours and the other two methods, we follow their default training strategy provided in their official implementation with five levels of detail and train NGLOD to convergence (for 250 epochs). Under this default setting, NGLOD uses 25 million sample points for training.
We summarize the recommended sample rate, its corresponding number of samples, training time in seconds, and the number of network parameters of each method on fitting \textit{Armadillo} in Tab.~\ref{tab:sample_number}.

\begin{table}[h]
    \centering
    \scalebox{0.8}{
    \begin{tabular}{c| c c c c }
    \hline
    &Sample rate & \#Samples $\downarrow$ & Train. time $\downarrow$ & Network param. $\downarrow$ \\
    \hline
    NGLOD~\cite{takikawa2021neural} & \# & 25.00M & 2,261 & 10.15M \\
    SPE~\cite{wang2021spline} & 156 & 1.54M & 1,810 & 2.14M \\
    SIREN~\cite{sitzmann2020implicit} & 126 & 816K & 1,502 & 1.58M \\ 
    Ours & 126 & 816K & 930 & 1.59M \\
    \hline
    \end{tabular}}
    \caption{\textbf{Statistics for different methods.} Training time in seconds.}
    \label{tab:sample_number}
\end{table}

\begin{figure}
    \centering
    \centering
    \begin{overpic}
    [width=\linewidth, clip]{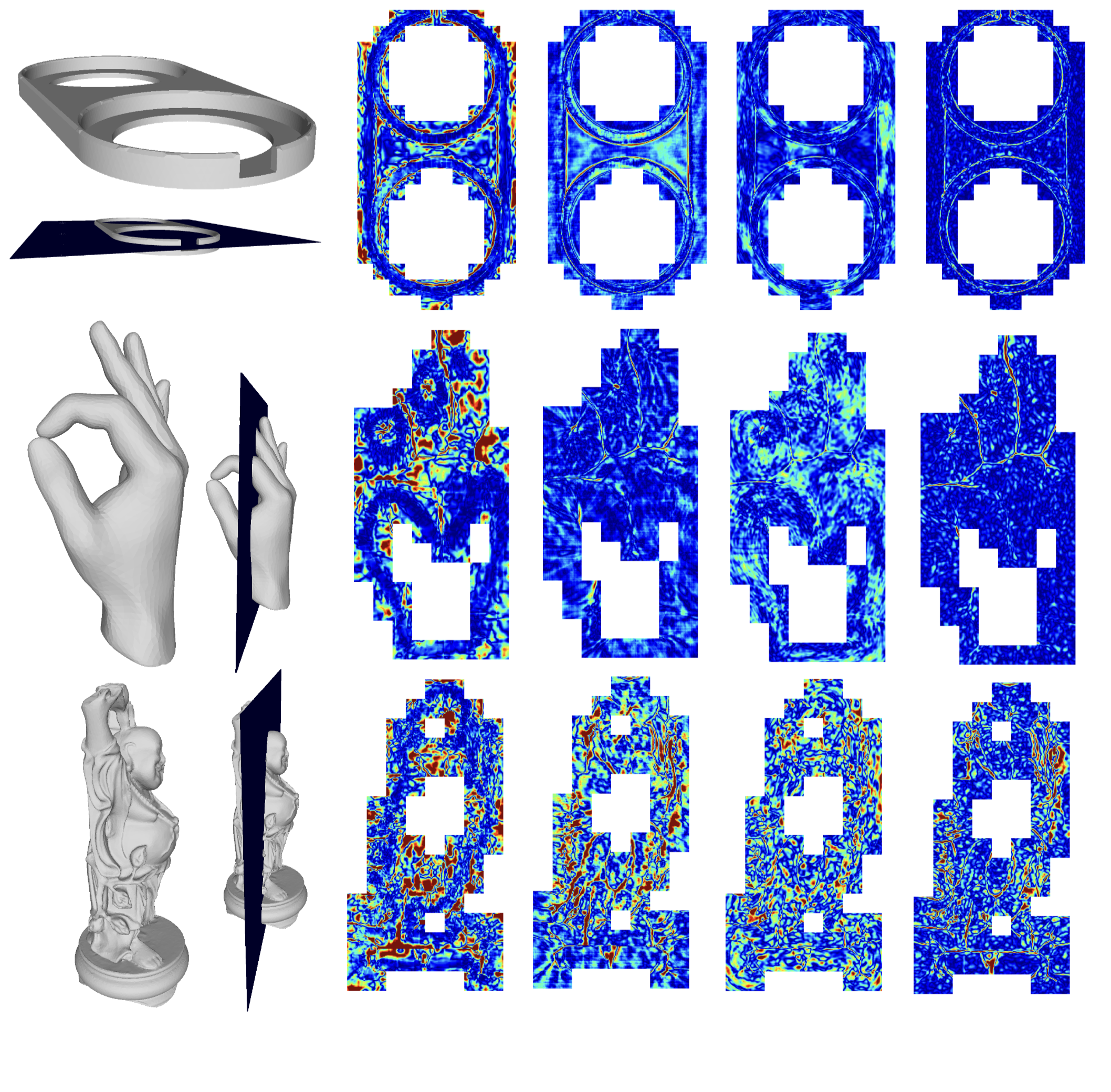}
    \put(10,3){(a) GT}
    \put(27,3){(b) NGLOD}
    \put(48,3){(c) SPE}
    \put(63,3){(d) SIREN}
    \put(83,3){(e) Ours}
    
    \end{overpic}
    \caption{\textbf{Comparison on SDF fitting task}. The first column shows the GT shapes and the cross-sectional planes. The SDF error at each point in the planes is shown. A warmer color indicates a higher error. The error value is clipped by a maximum of 0.003. }
    \label{fig:sdf_contour}
\end{figure}

The results in Table~\ref{tab:spatial_sota} show that our method outperforms the three baseline approaches in the SDF fitting task on more shapes. Some qualitative results are shown in Fig.~\ref{fig:sdf_contour}. The qualitative results further confirm that the amount of training samples suggested by our method enables accurate SDF fitting with a simple PE-equipped MLP network. 

\begin{table}[t]
    \centering
    \scalebox{0.9}{
    \begin{tabular}{c | c c c c  }
    \hline
        & Ours & SIREN~\cite{sitzmann2020implicit} & SPE~\cite{wang2021spline} & NGLOD~\cite{takikawa2021neural} \\
        \hline
        Fandisk & \textbf{4.10E-4} & 6.58E-4 & 6.29E-4 & 7.49E-4 \\
        Lucy & \textbf{8.68E-4} & 8.69E-4 & 9.22E-4  & 11.2E-4 \\
        Happy Buddha & \textbf{6.72E-4} & 6.75E-4 & 9.59E-4 & 11.3E-4 \\
        Hand  & \textbf{3.78E-4} & 6.50E-4 & 6.15E-4 & 7.96E-4 \\
        Bimba & \textbf{5.63E-4} & 5.95E-4 & 9.23E-4  & 8.54E-4 \\
        ABC1 & \textbf{3.18E-4} & 4.39E-4 & 5.64E-4 & 5.91E-4 \\
        ABC2 & \textbf{2.65E-4} & 4.73E-4 & 8.27E-4 & 7.03E-4 \\
        Armadillo & \textbf{6.83E-4} & 7.25E-4 & 6.74E-4 & 11.19E-4 \\ 
        Woman & \textbf{3.71E-4} & 5.93E-4 & 5.22E-4 & 11.01E-4\\
        
        \hline
    \end{tabular}}
    \caption{\textbf{Comparison on the SDF fitting task.} Reported are mean absolute SDF errors between the fitted SDF and the corresponding ground-truth SDF on evenly sampled validation points. Our PE-MLP networks with the suggested sufficient sampling rate outperform other comparing baselines by a large margin. 
    }
    \label{tab:spatial_sota}
\end{table}

\begin{figure*}
    \centering
    \centering
    \begin{overpic}
    [width=\linewidth, clip]{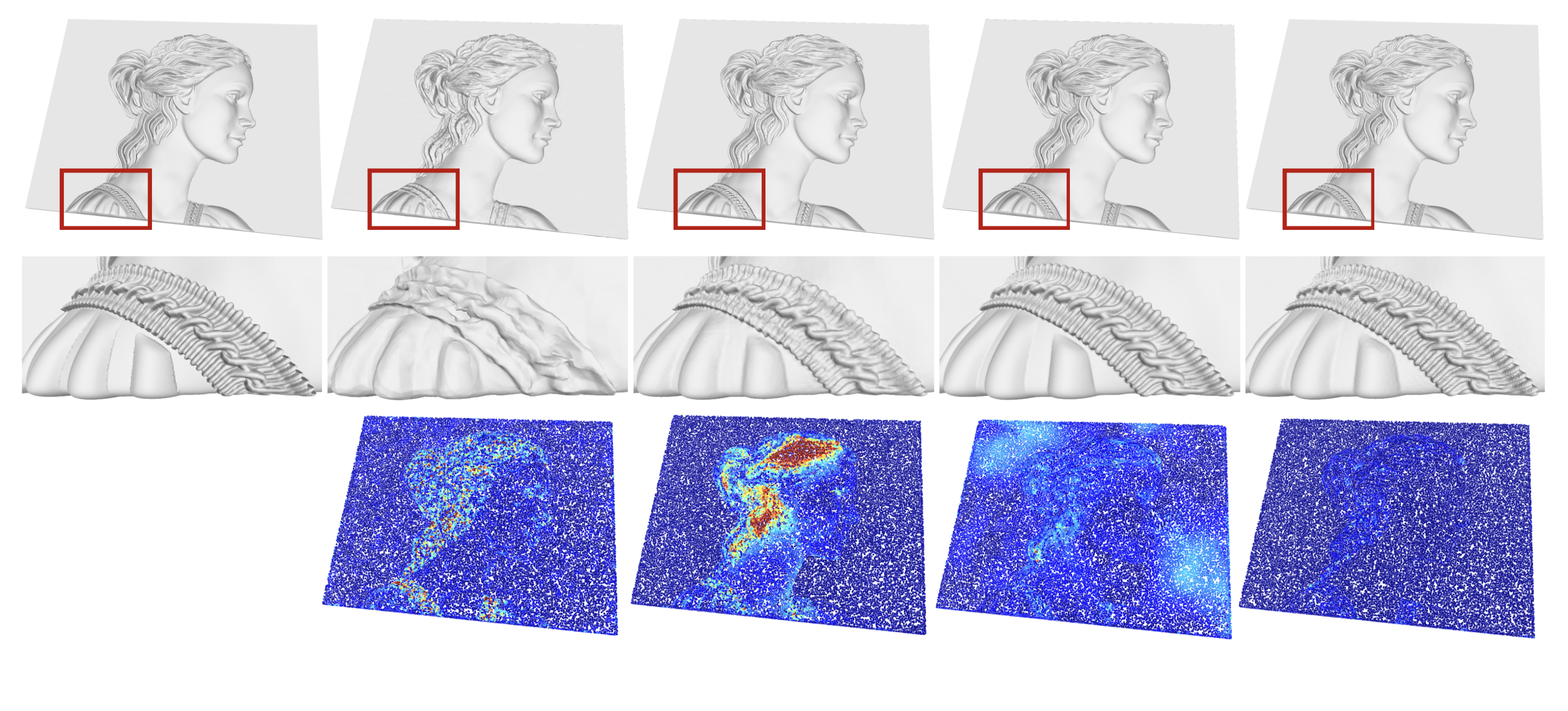}
    \put(8,3){(a) GT}
    \put(25,3){(b) NGLOD}
    \put(45,3){(c) SPE}
    \put(65,3){(d) SIREN}
    \put(85,3){(e) Ours}
    \end{overpic}
    \caption{
    \textbf{Qualitative comparison of surface fitting results for the Woman model}. The first line shows the extracted geometry, the second line provides a close-up view, and the third line displays error maps (red/blue indicates higher/lower error). NGLOD and SPE fail to capture high-frequency details, while SIREN exhibits wavy artifacts in flat areas. Our method excels in recovering high-frequency details and modeling flat regions, outperforming the baselines.
    }
    \label{fig:surface_geo}
\end{figure*}

\begin{figure}
    \centering
    \centering
    \begin{overpic}
    [width=\linewidth, clip]{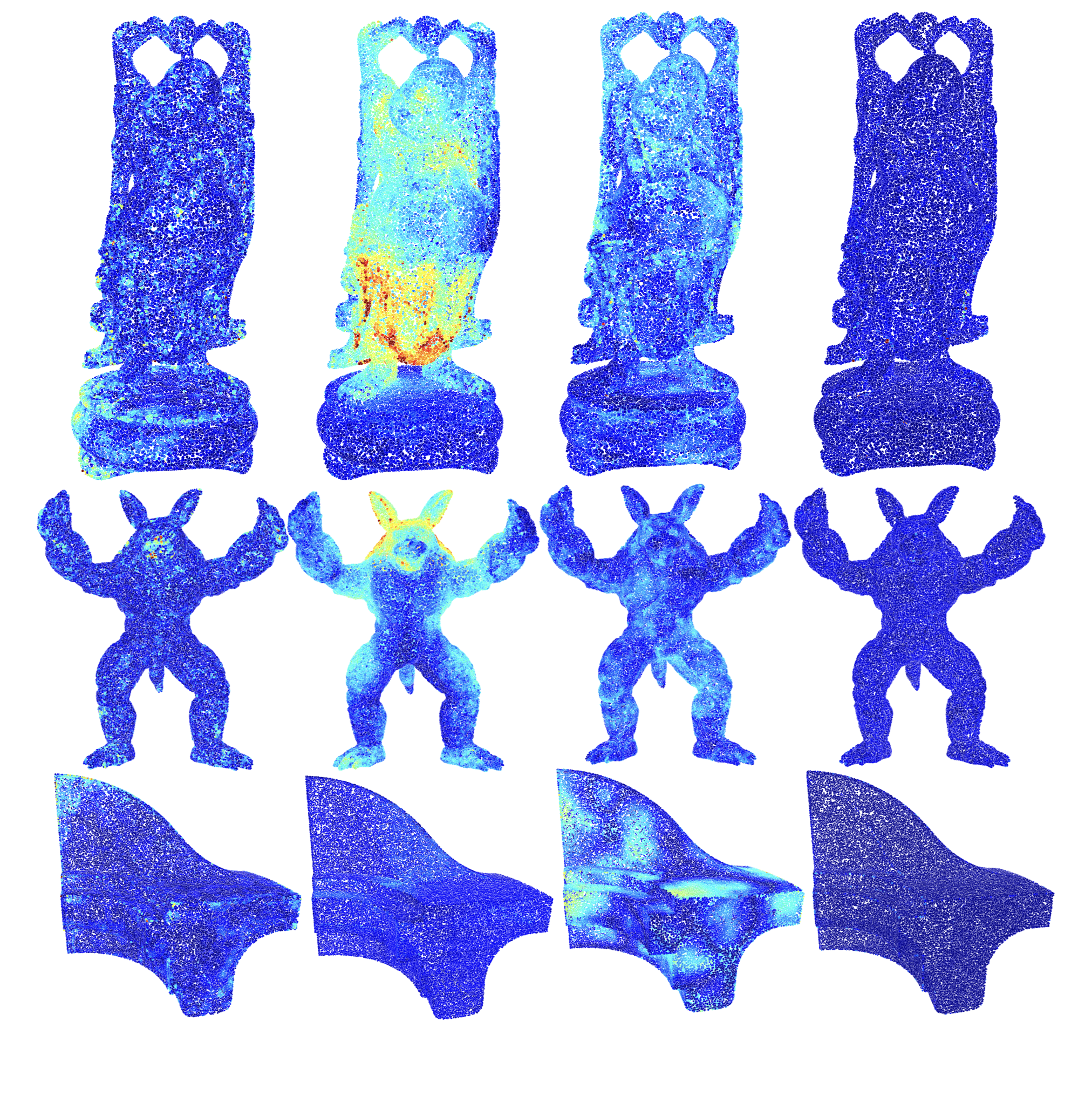}
    \put(6,2){(a) NGLOD}
    \put(33,2){(b) SPE}
    \put(52,2){(c) SIREN}
    \put(76,2){(d) Ours}
    \end{overpic}
    \caption{\textbf{Qualitative comparison on the surface fitting task}. The error maps show Chamfer Distances between the extracted surfaces and the ground truth (GT) meshes, clipped to a maximum of 0.003. Red indicates high error, while blue indicates low error.}
    \label{fig:surface_err_map}
\end{figure}

\begin{figure}
    \centering
    \begin{overpic}
    [width=\linewidth, clip]{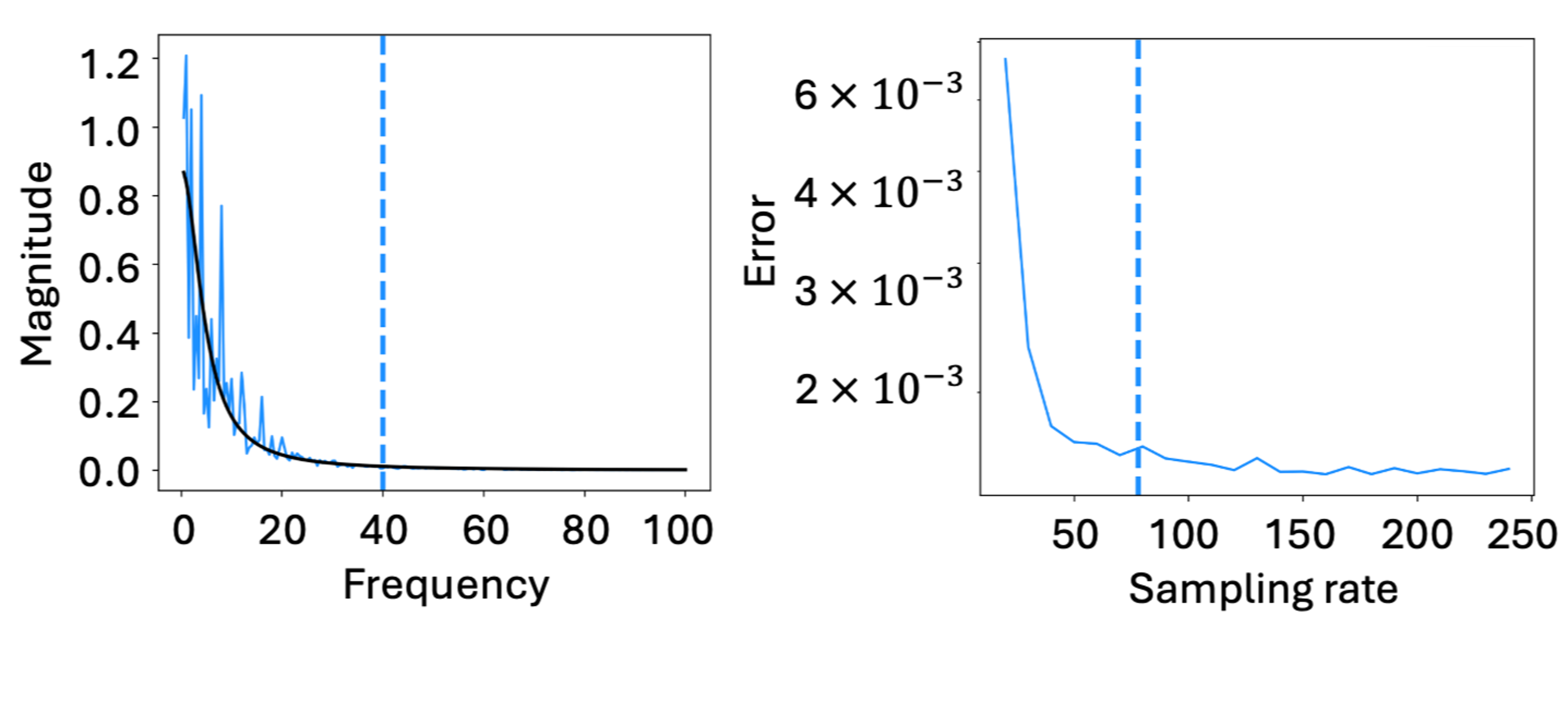}
    \put(13,1){(a) Spectrum }
    \put(63,1){(b) SDF Error}
    \end{overpic}
    \caption{
    \textbf{Smaller network.} The spectrum (a) and corresponding SDF error curve (b) of a PE-equipped MLP (4 layers, 128 neurons per layer, with $D=4$). Our recommended sampling rate allows this smaller PE-equipped MLP to achieve near-saturation SDF fitting accuracy.
    }
    \label{fig:sample_error_small}
\end{figure}

\begin{figure}
    \centering
    \begin{overpic}
    [width=\linewidth, clip]{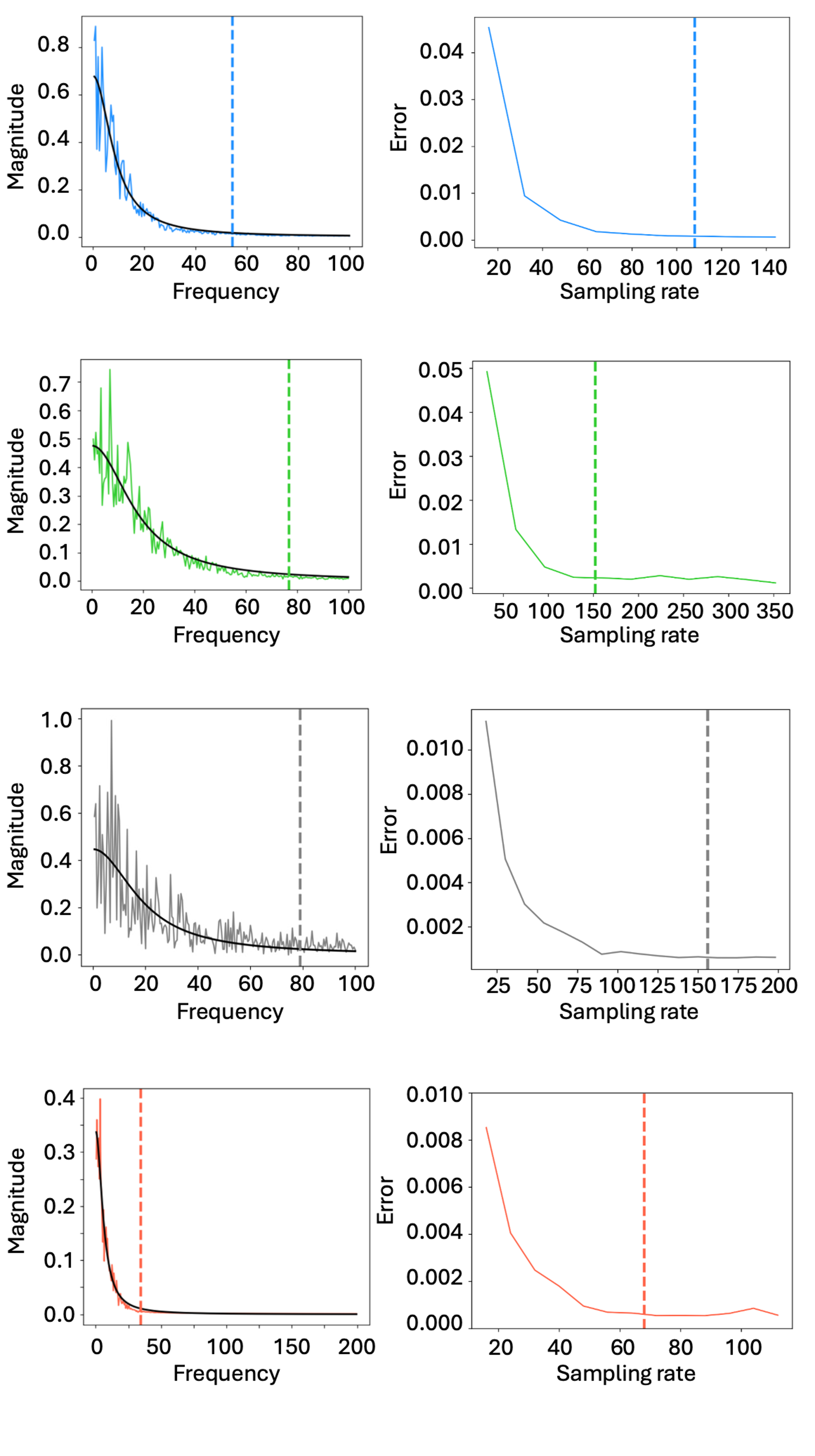}
    \put(22,77){(a) FPE ($\sigma$=10)}
    \put(22,53.5){(b) FPE ($\sigma$=19)}
    \put(22,28){(c) SPE}
    \put(22,2){(d) SIREN}
    \end{overpic}
    \caption{
    \textbf{Other network architectures.} Our sampling rule applies to FPE-MLP~\cite{tancik2020fourier} (a, b), {SPE}~\cite{wang2021spline}, and SIREN~\cite{sitzmann2020implicit} (c). Colored curves represent the intrinsic spectra, while the black curves show fitted approximations of the spectra. The vertical dashed lines on the right indicate the sampling rate corresponding to the cut-off frequency on the left, capturing the convergence of each network relative to the sampling rate.}
    \label{fig:siren_fpe}
    \vspace{-2mm}
\end{figure}

\begin{figure}
    \centering
    \centering
    \begin{overpic}
    [width=\linewidth, clip]{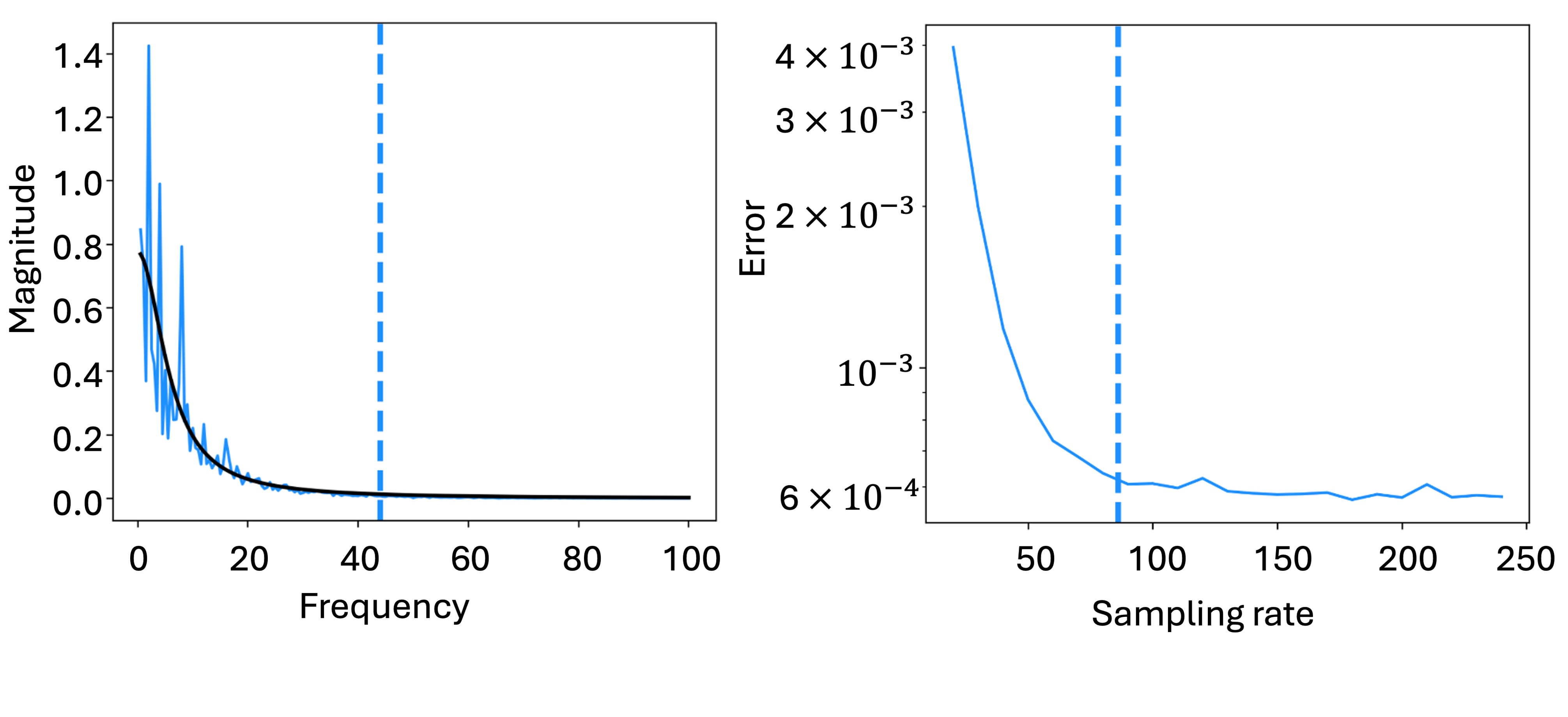}    
    \put(15,1){(a) Spectrum}
    \put(63,1){(b) SDF Error}
    \end{overpic}
    \caption{
    Xavier uniform initialization. The spectrum of a PE-equipped MLP ($D=4$) with Xavier initialization (a) and the corresponding SDF error curve (b). Dashed lines indicate the cut-off frequency (a) and corresponding sampling rate (b) which captures the convergence point of the fitting accuracy relative to the sampling rate.}
    \label{fig:xavier_init}
\end{figure}

\subsection{Discussions}
\label{sec:analysis}

In the following discussion, we provide additional analysis of the surface fitting quality, validation of our sampling distribution, the relationship between shape frequency and network complexity, and the relationship between the network intrinsic frequency and the size of trainable parameters.

\textbf{1) Can our method produce high-quality surfaces?}
To show that our method also enables accurate surface fitting, we further conduct a surface fitting experiment and compare our performance with the same baseline methods. In this experiment, we incorporate 10 million surface samples on the target shape in training our method and the baseline methods, which is the same as \cite{sitzmann2020implicit}. 

We report the quality of surface fitting results using the Chamfer Distance (CD) in Table~\ref{tab:surface_sota}. The results show that our method also outperforms all baseline methods on the surface fitting task. To provide additional visual evidence, we present a set of surfaces extracted from the learned SDF in Fig.~\ref{fig:surface_geo}. These visualizations highlight the ability of our method to capture high-frequency details while reproducing smooth, low-frequency areas. In comparison, while being able to represent geometric details to some extent, the baseline methods fail to accurately depict intricate features, as demonstrated in the zoomed-in views of Fig.~\ref{fig:surface_geo} (the second row). Furthermore, as depicted in the third row of Fig.~\ref{fig:surface_geo}, SIREN exhibits noticeable errors in the flat area. More comparison results are shown in Fig.~\ref{fig:surface_err_map}.

\begin{table}[t]
    \centering
    \scalebox{0.98}{
    \begin{tabular}{c| c c c c }
    \hline
    & Ours & SIREN~\cite{sitzmann2020implicit} & SPE~\cite{wang2021spline} & NGLOD~\cite{takikawa2021neural} \\
    \hline
    Fandisk & \textbf{2.29E-5} & 55.3E-5 & 24.2E-5 & 25.7E-5\\
    Lucy  & \textbf{7.90E-5} & 92.5E-5  & 102.4E-5 & 45.5E-5 \\ 
    Happy Buddha & \textbf{9.18E-5} & 56.1E-5   & 92.2E-5  & 41.9E-5\\
    Hand  & \textbf{9.48E-5} & 67.3E-5 & 32.1E-5 & 21.1E-5\\
    Bimba & \textbf{6.24E-5} & 46.4E-5 & 66.8E-5  & 31.5E-5\\ 
    ABC1 & \textbf{4.49E-5} & 62.6E-5 & 30.3E-5 & 24.2E-5 \\
    ABC2 & \textbf{2.39E-5} & 76.1E-5 & 15.1E-5 & 25.6E-5\\
    Armadillo & \textbf{17.4E-5} & 49.7E-5 & 92.3E-5 & 37.7E-5 \\ 
    Woman & \textbf{5.20E-5} & 35.0E-5 & 36.5E-5 & 25.3E-5 \\
    \hline
    \end{tabular}}
    \caption{\textbf{Comparison on the surface fitting task.} The Chamfer Distances (CD) between the extracted zero-level-set surfaces and ground-truth ones are reported.
    }
    \label{tab:surface_sota}
    \vspace{-2mm}
\end{table}

\textbf{2) Why not choose on-the-fly sampling?}
Given that we have established the requirement for the sampling rate to meet the lower bound set by the network's initial intrinsic frequency, it is natural to question why we do not opt for an on-the-fly sampling strategy similar to that initially employed in NGLOD\cite{takikawa2021neural}. The primary reason is that, in the task of SDF fitting, performing on-the-fly sampling at every iteration would be time-consuming, even with a GPU-accelerated algorithm. For instance, if we were to resample the training points at each iteration, the time spent on sampling would be 2.5 times longer than the training time in the case of NGLOD~\cite{takikawa2021neural}. 

While it is possible to decrease the sampling time by adopting interval-based sampling, as exemplified in the current implementation of NGLOD~\cite{takikawa2021neural}, the question arises: What is the appropriate duration for each interval? To address this inquiry, we must revisit the challenge of determining the optimal quantity of sampling points for effective training. Resolving this matter constitutes the primary emphasis of the present paper.

\textbf{3) Applicability to other settings} 
\textit{a) Smaller networks.}
As shown in Fig.~\ref{fig:sample_error_small}, to demonstrate that the sampling rate suggested by our approach is also applicable to smaller networks other than the default 8-layer MLP, we apply the same approach to analyzing the frequency of a smaller MLP network and conduct the same SDF fitting experiment using different sampling rates. This smaller network has 4 hidden layers with 128 neurons per layer. We observed the same trend as observed in the experiments in Sec.~\ref{sec:exp_sample_rate} that SDF fitting errors converge as the sampling rate increases beyond the suggested rate based on the cut-off frequency determined by our approach.

\textit{b) Other network architectures.} 
To examine to what extent the proposed sampling scheme is effective, we further apply the sampling scheme to some other MLP network architectures, i.e., FPE-MLP~\cite{tancik2020fourier}, {SPE}~\cite{wang2021spline}, and SIREN~\cite{sitzmann2020implicit}. As shown in Fig.~\ref{fig:siren_fpe}, the SDF fitting errors decrease and converge as the sampling rate reaches the proposed sampling rate, which demonstrates that the proposed intrinsic spectrum and cut-off frequency are also applicable to these architectures. 

\textit{c) Different initialization methods.} 
So far, we have seen that our approach applies to different network settings. 
One may be curious about whether our proposed approach is still applicable when the initialization scheme of the weights and biases of a network changes. Hence, as an example, we use Xavier uniform initialization~\cite{xavier} to replace the default uniform initialization used in previous experiments. Fig.~\ref{fig:xavier_init} empirically verifies the effectiveness of our approach in determining a cut-off frequency for preparing training samples.

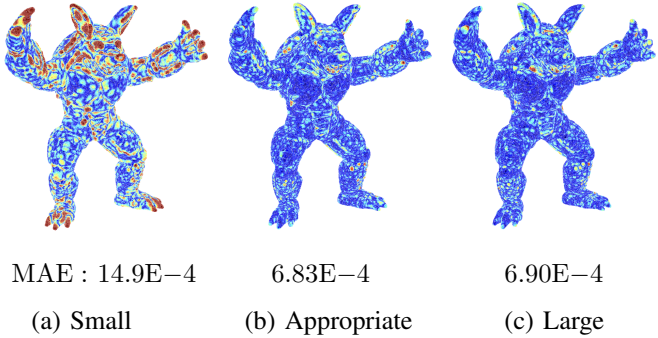
\begin{figure}
    \centering
    \begin{overpic}
    [width=\linewidth, trim = 100 0 0 0, clip]{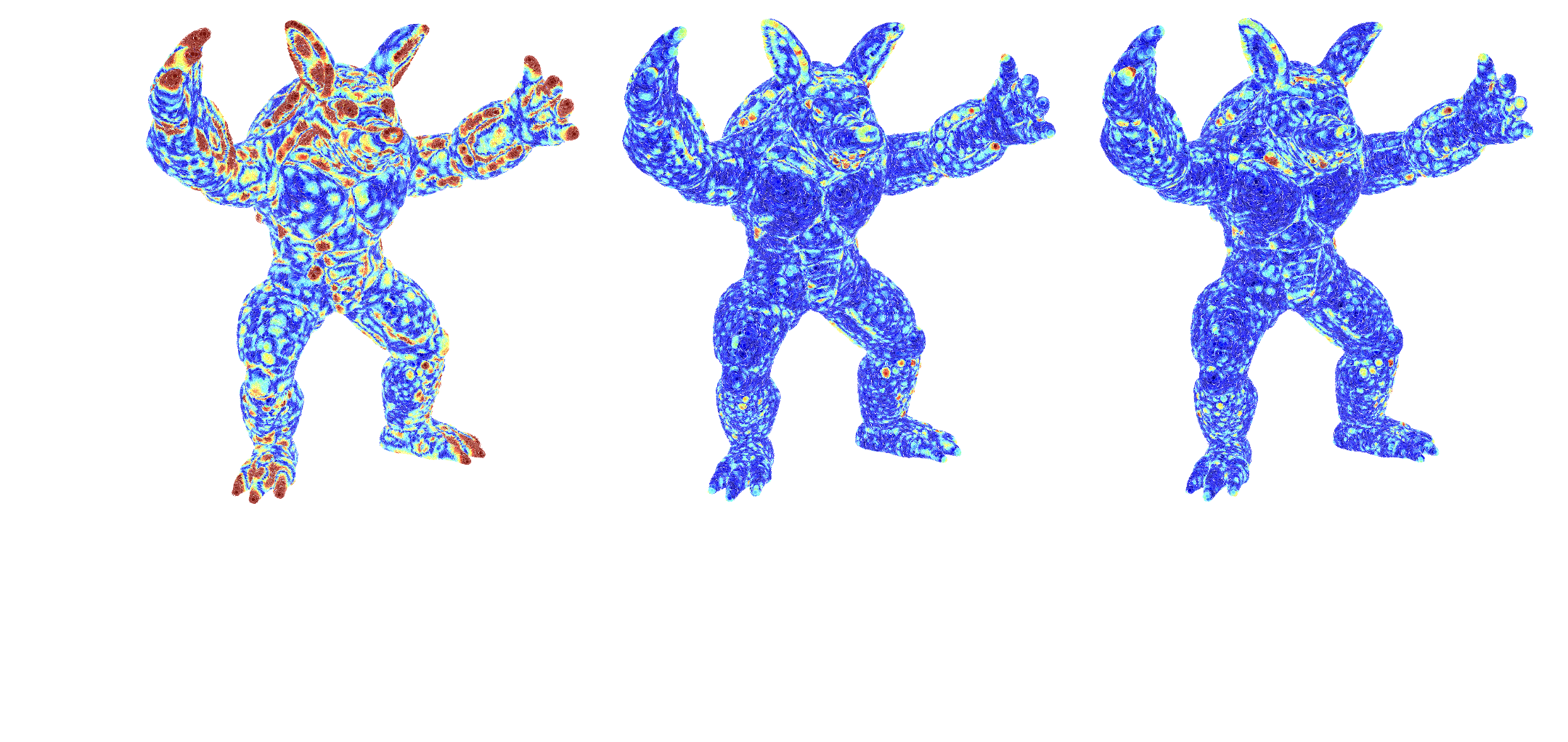}
    \put(1,9){$\mathrm{MAE:}~14.9\mathrm{E}{-4}$}
    \put(40,9){$6.83\mathrm{E}{-4}$}
    \put(75,9){$6.90\mathrm{E}{-4}$}
    \put(4,2){(a) Small}
    \put(36,2){(b) Appropriate}
    \put(75,2){(c) Large}
    \end{overpic}
    \caption{\textbf{The MLP network's capability for high-quality fitting.} The recommended sampling rate is provided for each PE-MLP of different sizes.
    Small PE-MLP has limited capacity for fitting a complex SDF. We visualize the SDF errors of surface points with a warmer color indicating a larger error. The mean absolute error (MAE) of SDF fitting is given.}
    \label{fig:diff_net}
    \vspace{-2mm}
\end{figure}

\textbf{4) Network capacity matters.} 
Though our method provides a lower bound for the sampling rate to avoid the aliasing effect for a given MLP network, an optimal sampling rate is not the only factor determining the final SDF fitting quality; attaining high-quality fitting results also requires a network with enough capacity to represent the given shape.
Fig.~\ref{fig:diff_net} shows the SDF fitting results of using networks of different capacities, all using the respective recommended sampling rates. Three PE-equipped MLP networks of different sizes are concerned: (a) \textit{Small} has 4 layers, each containing 128 neurons; (b) \textit{Appropriate} has 8 layers, each containing 512 neurons; (c) \textit{Appropriate} has 10 layers, each containing 512 neurons. 
As one can see, the small network's capacity proved to be inadequate for capturing the frequency of the target shape, resulting in a significant fitting error. 
The result produced by Fig.~\ref{fig:diff_net}(b) shows a very low level of error, highlighting the importance of using a sufficient sampling rate along with a network that has adequate capacity. Adding more layers to the network does not yield a noticeable improvement in accuracy, as shown in Fig.~\ref{fig:diff_net}(c).

\begin{figure}
    \centering
    \begin{overpic}
    [width=1.0\linewidth, clip]{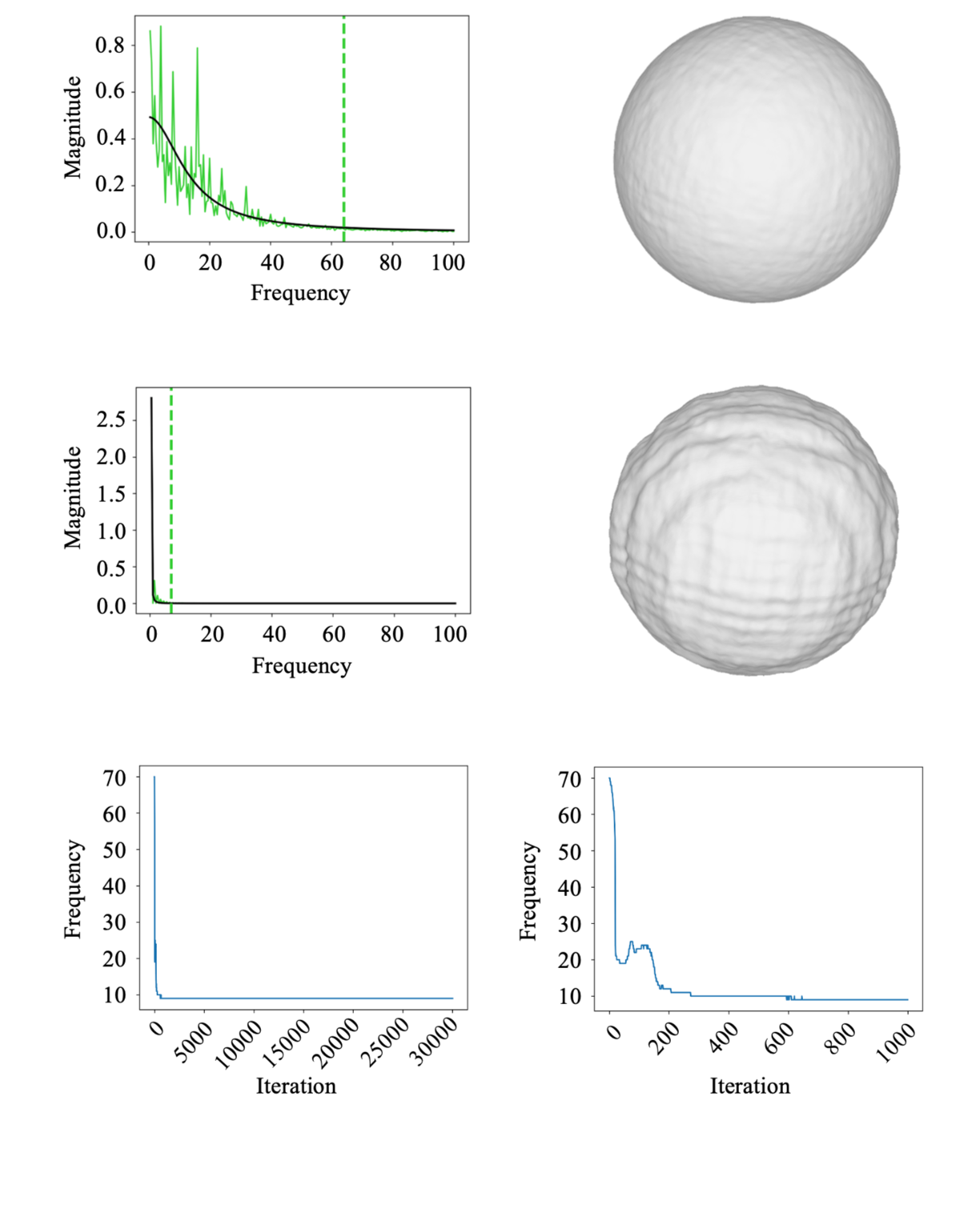}
    \put(23,72){(a)}
    \put(60,72){(b)}
    \put(23,41){(c)}
    \put(60,41){(d)}
    \put(23,7){(e)}
    \put(60,7){(f)}
    \end{overpic}
    \caption{\textbf{Frequency spectrum from randomized neural networks is intrinsic.} (a) Intrinsic frequency spectrum of several networks initialized randomly; (b) Surface extracted from the fitting results based on the recommended sampling rate corresponding to the cut-off frequency of (a); (c) Frequency spectrum of the trained neural network of (b); (d) Wavy artifacts observed when training the network on the sampling rate of cut-off frequency of (c).
    (e) shows the change of the network's cut-off frequency during the complete training process (30,000 iterations) with (f) zooming in to show the first 1,000 iterations. 
    }
    \label{fig:diff_init}
    \vspace{-2mm}
\end{figure}

\textbf{5) Sampling rate required for a PE-MLP initialized with a pretrained weight.} 
While the optimal sampling rate may indeed change as MLPs are trained progressively and converge to the target signal, the sampling rate determined based on the cut-off frequency of a randomly initialized network is sufficient for training the network as its frequency decreases. This ensures that wavy artifacts can be avoided even during progressive training. 
An example is provided where the network's spectra before and after training are shown in Fig. \ref{fig:diff_init}(a) and (c), respectively. The difference in the spectra leads to a noticeable decrease in the cut-off frequency (vertical dashed lines). Fig. \ref{fig:diff_init}(b) shows the high-quality result obtained by training the network from scratch with the sampling rate based on the cut-off frequency in Fig. \ref{fig:diff_init}(a). However, when we initialize the same network with the weights obtained in Fig. \ref{fig:diff_init}(c) and train this network with the sampling rate corresponding to the cut-off frequency in Fig. \ref{fig:diff_init}(c), artifacts are observed in the result in Fig. \ref{fig:diff_init}(d). 
Fig. \ref{fig:diff_init}(e) and (f) show the trend that the cut-off frequency decreases as the randomly initialized network converges to fit the target surface during the training process.

\begin{figure}[h]
    \centering
    \begin{overpic}
    [width=0.9\linewidth, clip]{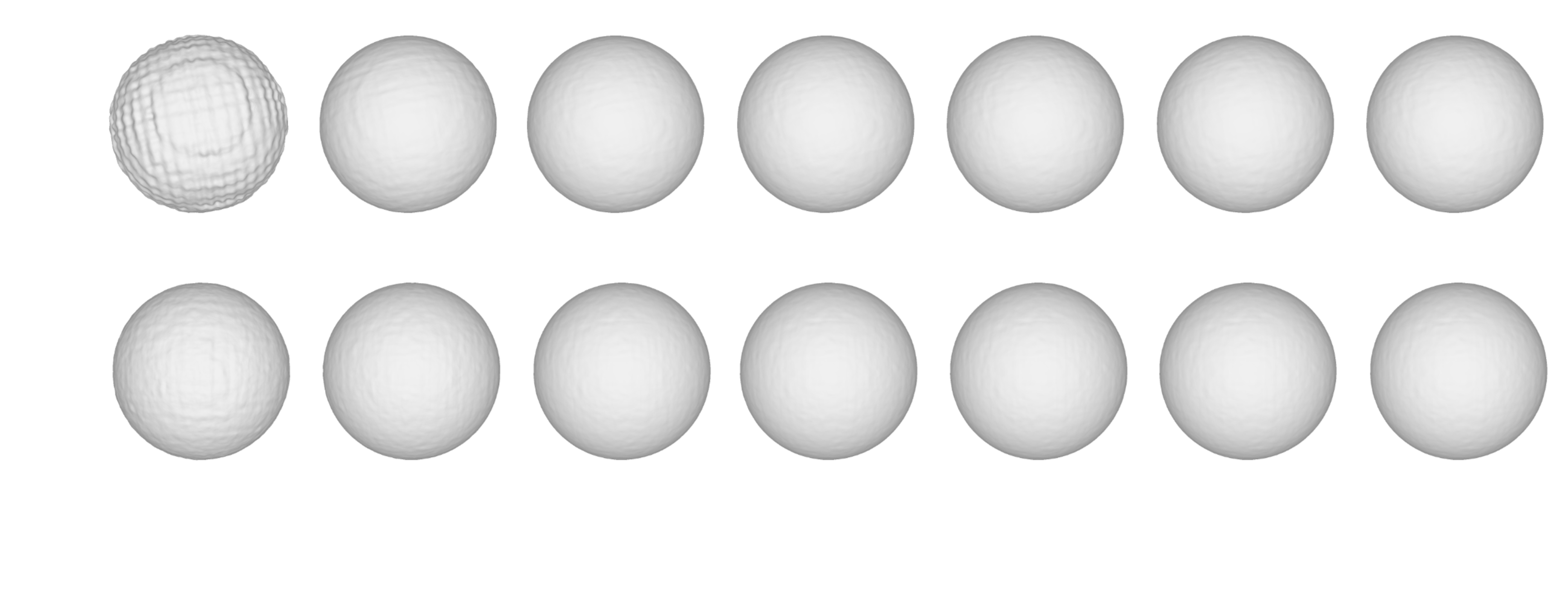}
    \put(-5, 32){LR}
    \put(-5, 28){$10^{-4}$}
    \put(-5, 17){LR}
    \put(-5, 13){$10^{-5}$}
    \put(-5, 5){SR}
    \put(12,5){15}
    \put(25,5){30}
    \put(39,5){45}
    \put(52,5){60}
    \put(65,5){75}
    \put(78,5){90}
    \put(91,5){105}
    \end{overpic}
    \caption{\\textbf{Higher sampling rate in need for preventing aliasing effect due to potentially high-frequency components during training.}
    Extracted isosurfaces from the PE-MLP network initialized by the pretrained weights but trained with different sampling rates (SR). Being low at initialization, the network's cut-off frequency may fluctuate due to the training dynamics as revealed by training the network with different learning rates (LR). 
    }
    \label{fig:diff_lr_sphere_init}
    \vspace{-2mm}
\end{figure}

To provide further insight into the above result, we conducted another experiment to show the sufficient sampling rate for the network initialized by the pretrained weights in Fig.~\ref{fig:diff_lr_sphere_init} where two different learning rates were used.
This example demonstrates that the cut-off frequency may fluctuate during the training process even if it is initially low. On the other hand, the cut-off frequency of randomly initialized networks is always stable and higher than the target signal, validating our approach and justifying our design choice based on the randomly initialized neural networks. 
Therefore, we recommend using the sampling rate corresponding to the cut-off frequency of randomly initialized networks to ensure the sampling rate is sufficient at any training stage.

\begin{figure}
    \centering
    \begin{overpic}
    [width=\linewidth, clip]{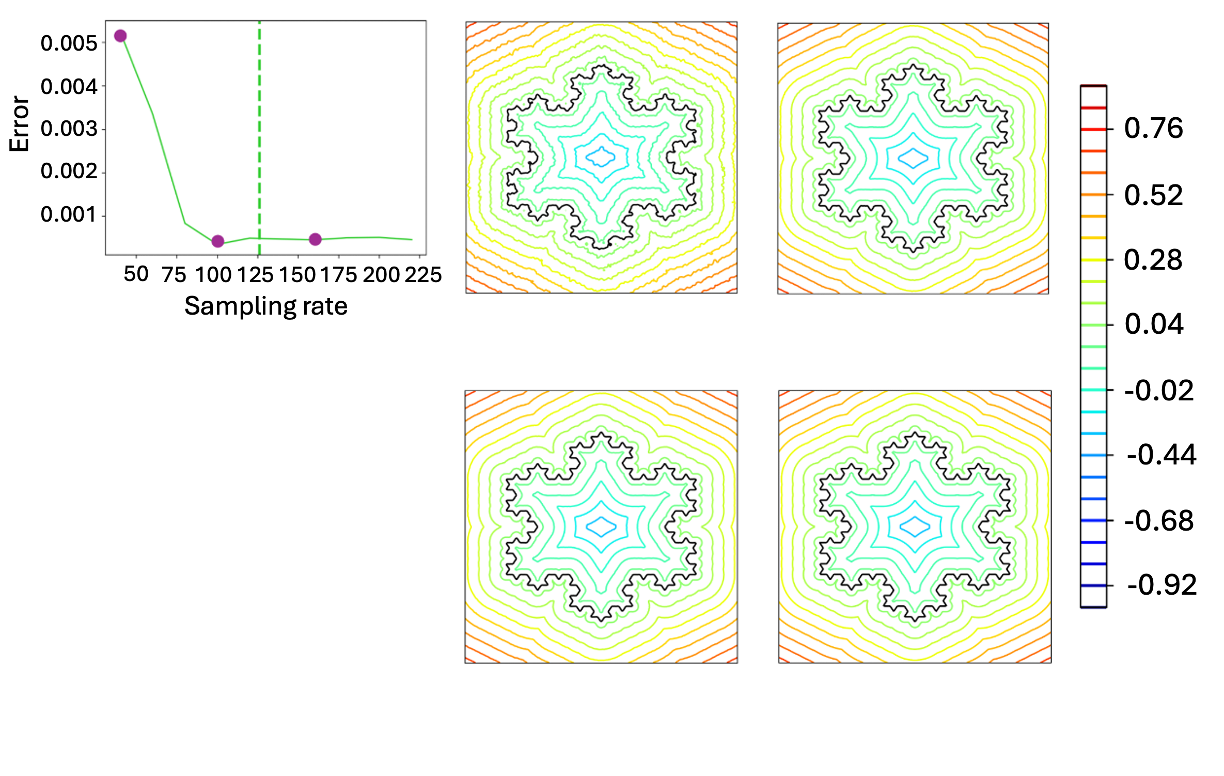}
    \put(9,33){(a) Error curve }
    \put(44,33){(b) 40}
    \put(70,33){(c) 100}
    \put(44,2){(d) 160}
    \put(64,2){(e) Ground truth}
    \end{overpic}
    \caption{
    \textbf{Representing vector graphics of Koch snowflake.} (a) Errors correspond to different sampling rates; the vertical dashed line indicates the sampling rate based on the network's cut-off frequency. Iso-contours of learned SDF and its zero-level set (black) at different sampling rates are shown in (b,c,d). Insufficient sampling rate (b) leads to noisy iso-contours.
    Ground-truth Koch snowflake at a maximum degree of 3 is depicted in (e). The color bar shows the distance value of the iso-contours.}
    \label{fig:snow_exp}
    \vspace{-2mm}
\end{figure}

\textbf{6) Application to represent vector graphics.} 
We performed an experiment with a \textit{Koch snowflake}, setting its maximum degree to 3. As shown in Fig.~\ref{fig:snow_exp}, while excessively low sampling frequencies introduce significant periodic noise in the iso-contours, an adequate sampling frequency recommended by our method can mitigate this noise and yield high-quality fitting results. 

\section{Conclusions}\label{sec:conclusion}
We investigated the role of spatial samples with ground-truth signed distance values in learning a quality signed distance field induced by the given shape and how the spatial samples can work in synergy with the sinusoidal positional encoding for this purpose. 
While it has been well known in a qualitative sense that more training samples can lead to better results of SDF approximation, we propose an efficient method for quantitatively estimating a sufficient sampling rate for training the neural network by analyzing its frequency spectrum and determining the cut-off frequency on the spectrum. We approach the problem by employing the Nyquist-Shannon to the intrinsic frequency of the PE-equipped MLP, which is attained by applying FFT analysis to the output of the PE-MLP with random initialization. We also demonstrate that by training with our recommended sample rate, the coordinate networks can achieve state-of-the-art performance regarding SDF accuracy. This provides a strong baseline for future studies in this field.

\textbf{Limitation and future work.} Although our approach finds its theoretical ground in the Nyquist-Shannon sampling theorem, the algorithm for determining the cut-off frequency is based on extensive experiments. Lack of theoretical proof of the cut-off frequency is a major limitation of our work. 

In the future, we aim to provide theoretical insights into the frequency spectrum of a deep neural network and the determination of its cut-off frequency for the fitting task. So far, we only focus on the MLP neural networks designed for fitting tasks. We are curious if this frequency-based analysis can be applied to other deep learning applications to better understand the inner working of the deep neural networks.

\bibliographystyle{IEEEtran}
\bibliography{egbib}

\section{Appendix}\label{sec:appendix}

\begin{figure*}
    \centering
    \begin{overpic}
    [width=\linewidth, clip]{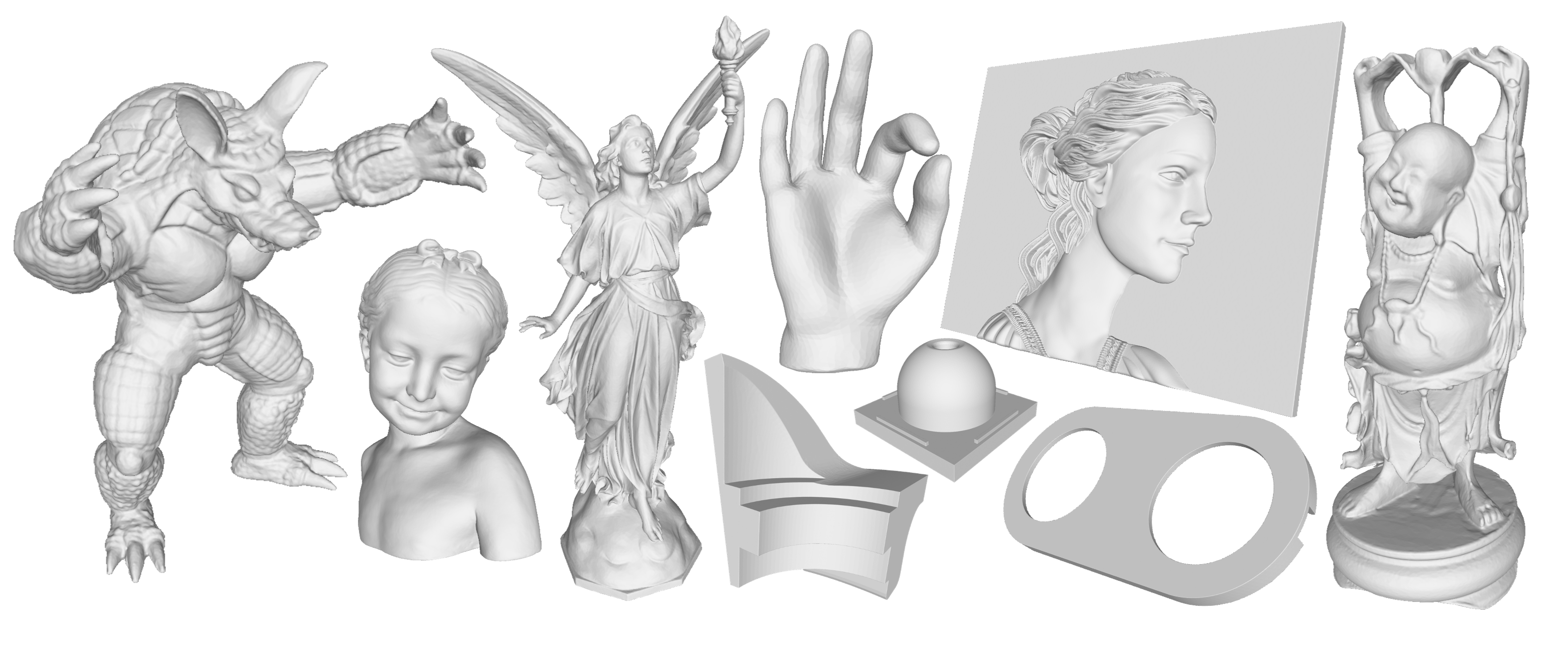}
    \end{overpic}
    \caption{\textbf{The fitted surfaces of our method in the surface fitting task.} These shapes contain various geometric details, such as flat regions, sharp features, high-frequency details, etc. 
    }
    \label{fig:gt_gallery}
    \vspace{-2mm}
\end{figure*}

\subsection{More Implementation Details}

\textbf{Dataset}. The dataset utilized in our experiments comprises 9 shapes that encompass a range of geometric details, as illustrated in Fig.~\ref{fig:gt_gallery}. Specifically, we obtained the models for \textit{Happy Buddha}\cite{curless1996volumetric}, \textit{Armadillo}, and \textit{Lucy} from the Stanford 3D Scanning Repository. The \textit{Fandisk} model was published in\cite{hoppe1994piecewise}. Both the \textit{Bimba} and \textit{Fandisk} models were retrieved from the following GitHub repository: https://github.com/alecjacobson/common-3d-test-models/. Additionally, we conducted tests using the \textit{Woman Relief} model, which was purchased online. Furthermore, we included two CAD models, namely \textit{ABC1} and \textit{ABC2}, from the ABC dataset~\cite{Koch2019ABC}. Lastly, the \textit{Hand} shape was generated by~\cite{hc2022arap}.

\subsection{More Experiments and Results}

In the following section, we present a series of supplementary experiments to validate our sampling rate, sampling distributions, and the extension from one dimension to three dimensions.

\begin{figure}
    \centering
    \begin{overpic}
    [width=\linewidth, clip]{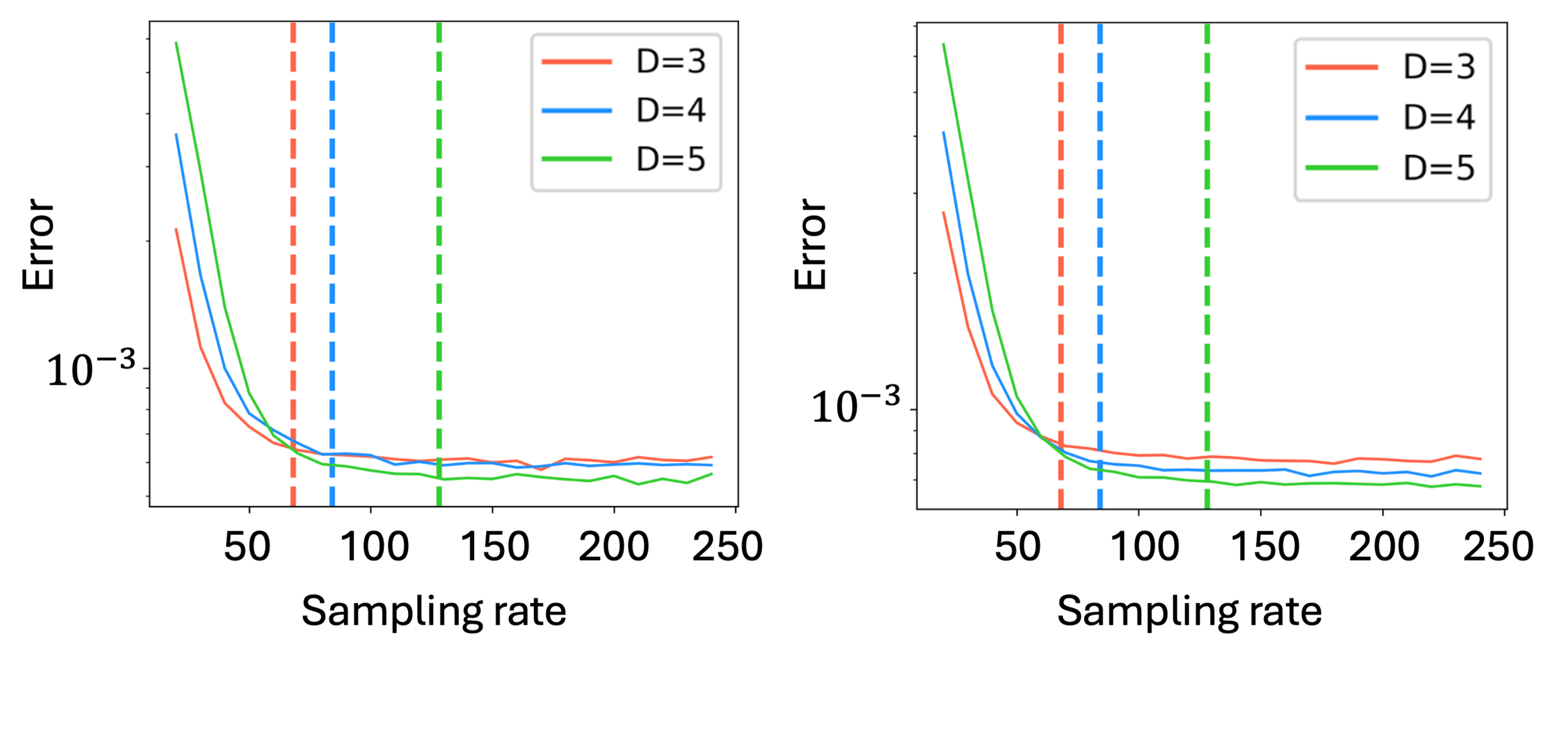}
    \put(20,2){(a) Bimba}
    \put(65,2){(b) Armadillo}
    \end{overpic}
    \caption{\textbf{Validation of our sampling rate on more shapes.} We show the SDF error curves of \textit{Bimba} and \textit{Armadillo} in (a) and (b). The vertical lines are our recommended sampling rates. They align well with the convergence points on the error curves.}
    \label{fig:more_shapes} 
    \vspace{-2mm}
\end{figure}

\textbf{Validation of our sampling rate on more shapes}.
We validate our sampling rate on additional shapes, namely \textit{Bimba} and \textit{Armadillo}. The SDF error curves for these shapes are illustrated in Fig.~\ref{fig:more_shapes}. It can be seen from the figure that our recommended sampling rates consistently correspond to the convergence points of the error curves, further confirming the superiority of our sampling method.

\begin{figure}
    \centering
    \begin{overpic}
    [width=\linewidth, clip]{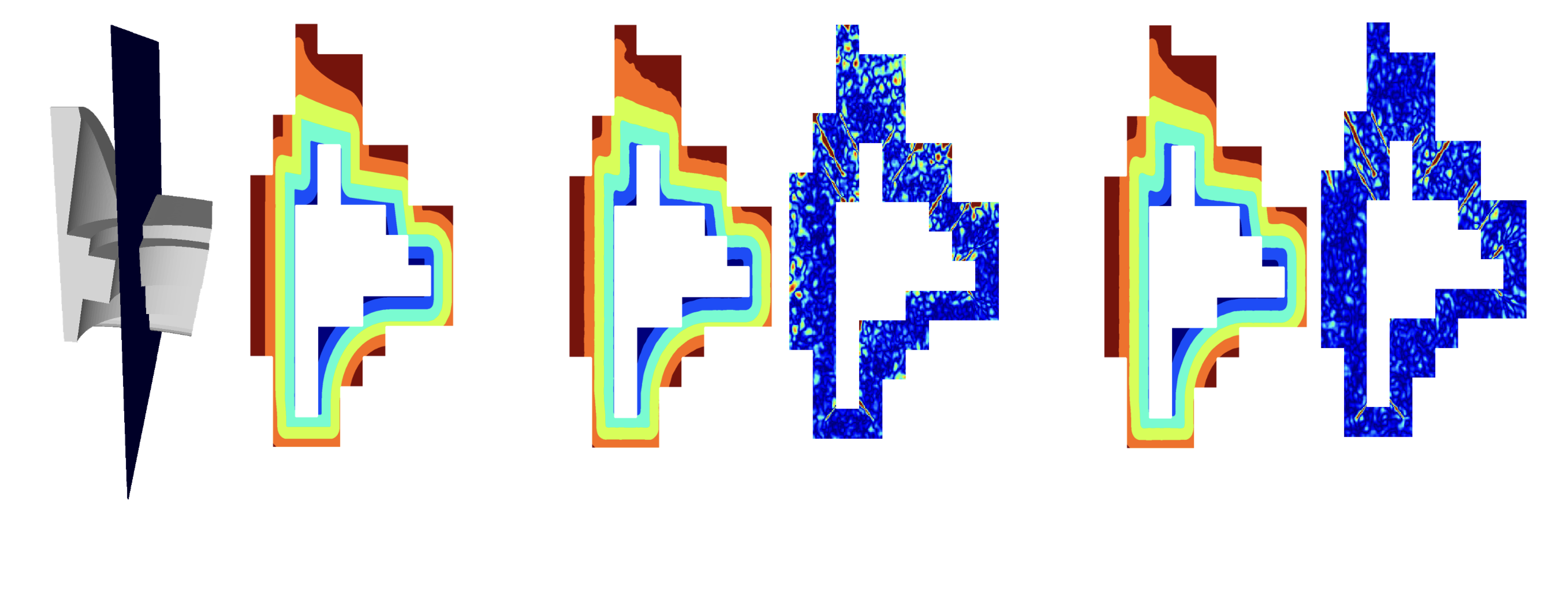}
    \put(10,4){(a) GT}
    \put(40,4){(b) Offset}
    \put(75,4){(c) Ours}
    \end{overpic}
    \caption{\textbf{Qualitative comparison of learned SDFs with different sampling schemes on \textit{Fandisk}.} In (b) and (c), the left column is the contour of the SDF slice as cut by the black plane in (a), and the right column is the corresponding error map of the SDF slice. The red indicates a higher error while the blue indicates a lower error. Our result (c) approximates the GT SDF (a) well compared with the \textit{Offset} scheme (b).}
    \label{fig:distribution}
    \vspace{-2mm}
\end{figure}

\begin{figure*}
    \centering
    \begin{overpic}
    [width=\linewidth, clip]{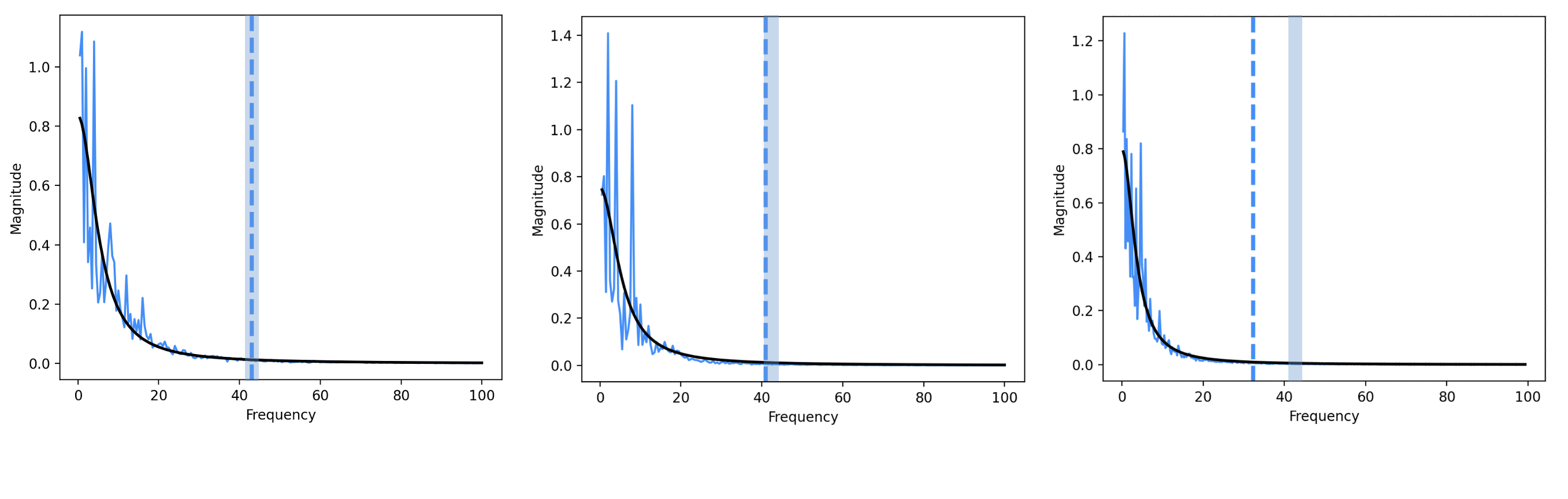}
    \put(10,2){(a) 1D Input }
    \put(40,2){(b) 3D Input on Axes}
    \put(73,2){(c) 3D Input on Diagonal}

    \end{overpic}
    \caption{\textbf{Extending our 1D FFT analysis to a PE-equipped MLP with 3D input.} We plot the spectrum of the output of the PE-equipped MLP with 3D input, where the points are sampled along the x-axis in (b), and the spectrum of the output of the PE-equipped MLP with 1D input in (a). Additionally, we display the spectrum of the output of the PE-equipped MLP with 3D input, where the points are sampled along the diagonal in (c). In order to better compare the cut-off frequencies of different spectra, we have marked the frequency 42
    with a vertical line of light blue color. The results indicate that the behavior of the spectrum of the PE-equipped MLP with 3D input along the axes is the same as that of the PE-equipped MLP with 1D input. The spectrum of the sample points along the diagonal exhibits significantly lower cut-off frequencies. The reason is that we adopt the axis-aligned positional encoding. Thus, we choose the frequency on axes as the reference for computing our sampling rate to ensure that it is able to cover the high frequency on the XYZ axes.}
    \label{fig:1d_3d_fft}
    \vspace{-2mm}
\end{figure*}

\textbf{Extension to PE-equipped MLP with 3D input}. We determine the cut-off frequency corresponding to the sampling rate by analyzing the frequency spectrum of the PE-MLP with a 1D input. However, when performing SDF fitting for a specific shape, our input becomes 3D. To show that the 3D frequency analysis is consistent to the 1D version, 
we sampled points along the x, y, z, and diagonal axes in the 3D domain and performed a 1D FFT analysis on the obtained signals. As depicted in Fig.\ref{fig:1d_3d_fft} (a) and (b), the cut-off frequency of a specific axis of the 3D input remains consistent with that of the 1D input, as shown in Fig.~\ref{fig:1d_3d_fft} (c) and (d). We also calculated the frequencies along the diagonal axis in the 3D domain. As illustrated in Fig.~\ref{fig:1d_3d_fft} (e) and (f), the frequencies along the diagonal axis are lower than those along the x, y, and z axes because we use the axis-aligned positional encoding. Hence, when determining the sampling density in the 3D domain, it is sufficient to use the frequencies along the x, y, and z axes to ensure it captures the highest frequency in space.

\begin{table}[t]
    \centering
    \scalebox{0.9}{
    \begin{tabular}{c|c c c c c   }
    \hline
        &  Fandisk & Lucy & Happy Buddha & Hand & Bimba \\ 
        \hline
        Ours & \textbf{4.10E-4} & \textbf{8.68E-4} & \textbf{6.72E-4} & \textbf{3.78E-4} & \textbf{5.63E-4} \\
        Offset &4.67E-4 & 9.26E-4 & 7.90E-4 & 5.62E-4 & 5.81E-4  \\
        \hline
    \end{tabular}}
    \caption{\textbf{To effectively control the high frequencies introduced by PE-MLP, uniform sampling is used.} We conducted a comparison of the SDF error between two sampling distributions. The first distribution is our uniform sampling, while the second, called ``Offset", is the Gaussian distribution~\cite{park2019deepsdf}. Both sampling distributions use an equal number of sampling points recommended by our sampling strategy. By utilizing uniform sampling, we can ensure stable control over the artifacts caused by PE-MLP, thereby recommending an optimal sampling number.}
    \label{tab:ablation_distribution}
    \vspace{-2mm}
\end{table}

\textbf{Is our uniform sampling a good sample distribution?}
In our implementation, we use uniform sampling in generating training examples as suggested by NS theorem. However, there is an alternative method~\cite{park2019deepsdf} where surface samples are perturbed using offset vectors generated from a Gaussian distribution. We refer to this sampling scheme and its corresponding results as \textit{Offset} in our experiments. From Tabble~\ref{tab:ablation_distribution}, we see that our sampling scheme consistently produces better results regarding the RMSE of SDF values than the \textit{Offset} scheme. Moreover, it is evident from Fig.~\ref{fig:distribution} that utilizing the \textit{Offset} sampling schema introduces wavy artifacts in regions where the SDF is distant from the surface. In contrast, our uniform distribution guarantees stability across the entire domain. This is because the high frequencies generated by PE-MLP are distributed throughout the domain. Consequently, even when a sufficient number of samples are present, we recommend utilizing a uniform distribution as a foundation to effectively cover the highest frequencies introduced by PE-MLP. Based on this, one can also apply additional emphasis on specific areas of interest.

\end{document}